# AND/OR Multi-Valued Decision Diagrams (AOMDDs) for Graphical Models

**Robert Mateescu**                                   MATEESCU@PARADISE.CALTECH.EDU
*Electrical Engineering Department*
*California Institute of Technology*
*Pasadena, CA 91125, USA*

**Rina Dechter**                                      DECHTER@ICS.UCI.EDU
*Donald Bren School of Information and Computer Science*
*University of California Irvine*
*Irvine, CA 92697, USA*

**Radu Marinescu**                                    R.MARINESCU@4C.UCC.IE
*Cork Constraint Computation Centre*
*University College Cork, Ireland*

## Abstract

Inspired by the recently introduced framework of AND/OR search spaces for graphical models, we propose to augment Multi-Valued Decision Diagrams (MDD) with AND nodes, in order to capture function decomposition structure and to extend these compiled data structures to general weighted graphical models (e.g., probabilistic models). We present the *AND/OR Multi-Valued Decision Diagram* (AOMDD) which compiles a graphical model into a canonical form that supports polynomial (e.g., solution counting, belief updating) or constant time (e.g. equivalence of graphical models) queries. We provide two algorithms for compiling the AOMDD of a graphical model. The first is search-based, and works by applying reduction rules to the trace of the memory intensive AND/OR search algorithm. The second is inference-based and uses a Bucket Elimination schedule to combine the AOMDDs of the input functions via the the APPLY operator. For both algorithms, the compilation time and the size of the AOMDD are, in the worst case, exponential in the *treewidth* of the graphical model, rather than *pathwidth* as is known for ordered binary decision diagrams (OBDDs). We introduce the concept of *semantic treewidth*, which helps explain why the size of a decision diagram is often much smaller than the worst case bound. We provide an experimental evaluation that demonstrates the potential of AOMDDs.

## 1. Introduction

The paper extends decision diagrams into AND/OR multi-valued decision diagrams (AOMDDs) and shows how graphical models can be compiled into these data-structures. The work presented in this paper is based on two existing frameworks: (1) AND/OR search spaces for graphical models and (2) decision diagrams.

### 1.1 AND/OR Search Spaces

AND/OR search spaces (Dechter & Mateescu, 2004a, 2004b, 2007) have proven to be a unifying framework for various classes of search algorithms for graphical models. The main characteristic is the exploitation of independencies between variables during search, which can provide exponential speedups over traditional search methods that can be viewed as traversing an OR structure. The AND nodes capture problem decomposition into *independent subproblems*, and the OR nodes represent branching according to variable values. AND/OR spaces can accommodate dynamic variable ordering, however most of the current work focuses on static decomposition. Examples of AND/OR search trees and graphs will appear later, for example in Figures 6 and 7.

The AND/OR search space idea was originally developed for heuristic search (Nilsson, 1980). In the context of graphical models, AND/OR search (Dechter & Mateescu, 2007) was also inspired by search advances introduced sporadically in the past three decades for constraint satisfaction and more recently for probabilistic inference and for optimization tasks. Specifically, it resembles the pseudo tree rearrangement (Freuder & Quinn, 1985, 1987), that was adapted subsequently for distributed constraint satisfaction by Collin, Dechter, and Katz (1991, 1999) and more recently by Modi, Shen, Tambe, and Yokoo (2005), and was also shown to be related to graph-based backjumping (Dechter, 1992). This work was extended by Bayardo and Miranker (1996) and Bayardo and Schrag (1997) and more recently applied to optimization tasks by Larrosa, Meseguer, and Sanchez (2002). Another version that can be viewed as exploring the AND/OR graphs was presented recently for constraint satisfaction (Terrioux & Jégou, 2003b) and for optimization (Terrioux & Jégou, 2003a). Similar principles were introduced recently for probabilistic inference, in algorithm Recursive Conditioning (Darwiche, 2001) as well as in Value Elimination (Bacchus, Dalmao, & Pitassi, 2003b, 2003a), and are currently at the core of the most advanced SAT solvers (Sang, Bacchus, Beame, Kautz, & Pitassi, 2004).

## 1.2 Decision Diagrams

Decision diagrams are widely used in many areas of research, especially in software and hardware verification (Clarke, Grumberg, & Peled, 1999; McMillan, 1993). A BDD represents a Boolean function by a directed acyclic graph with two terminal nodes (labeled 0 and 1), and every internal node is labeled with a variable and has exactly two children: *low* for 0 and *high* for 1. If isomorphic nodes were not merged, we would have the full search *tree*, also called Shannon tree, which is the usual full tree explored by a backtracking algorithm. The tree is ordered if variables are encountered in the same order along every branch. It can then be compressed by merging isomorphic nodes (i.e., with the same label and identical children), and by eliminating redundant nodes (i.e., whose *low* and *high* children are identical). The result is the celebrated *reduced ordered binary decision diagram*, or OBDD for short, introduced by Bryant (1986). However, the underlying structure is OR, because the initial Shannon tree is an OR tree. If AND/OR search trees are reduced by node merging and redundant nodes elimination we get a compact search graph that can be viewed as a BDD representation augmented with AND nodes.

## 1.3 Knowledge Compilation for Graphical Models

In this paper we combine the two ideas, creating a decision diagram that has an AND/OR structure, thus exploiting problem decomposition. As a detail, the number of values is also increased from two to any constant. In the context of constraint networks, decision diagrams can be used to represent the whole set of solutions, facilitating solutions count, solution enumeration and queries on equivalence of constraint networks. The benefit of moving from OR structure to AND/OR is in a lower complexity of the algorithms and size of the compiled structure. It typically moves from being bounded exponentially in *pathwidth $pw^*$*, which is characteristic to chain decompositions or linear structures, to being exponentially bounded in *treewidth $w^*$*, which is characteristic of tree structures (Bodlaender & Gilbert, 1991) (it always holds that $w^* \leq pw^*$ and $pw^* \leq w^* \cdot \log n$, where $n$ is the number of variables of the model). In both cases, the compactness result achieved in practice is often far smaller than what the bounds suggest.

A decision diagram offers a compilation of a propositional knowledge-base. An extension of the OBDDs was provided by Algebraic Decision Diagrams (ADD) (Bahar, Frohm, Gaona, Hachtel, Macii, Pardo, & Somenzi, 1993), where the terminal nodes are not just 0 or 1, but take values from an arbitrary finite domain. The *knowledge compilation* approach has become an important research direction in automated reasoning in the past decade (Selman & Kautz, 1996; Darwiche & Marquis, 2002; Cadoli & Donini, 1997). Typically, a knowledge representation language is compiled into a compact data structure that allows fast responses to various queries. Accordingly, the computational effort can be divided between an *offline* and an *online* phase where most of the work is pushed offline. Compilation can also be used to generate compact building blocks to be used by online algorithms multiple times. Macro-operators compiled during or prior to search can be viewed in this light (Korf & Felner, 2002), while in graphical models the building blocks are the functions whose compact compiled representations can be used effectively across many tasks.

As one example, consider product configuration tasks and imagine a user that chooses sequential options to configure a product. In a naive system, the user would be allowed to choose any valid option at the current level based only on the initial constraints, until either the product is configured, or else, when a dead-end is encountered, the system would backtrack to some previous state and continue from there. This would in fact be a search through the space of possible partial configurations. Needless to say, it would be very unpractical, and would offer the user no guarantee of finishing in a limited time. A system based on compilation would actually build the *backtrack-free* search space in the offline phase, and represent it in a compact manner. In the online phase, only valid partial configurations (i.e., that can be extended to a full valid configuration) are allowed, and depending on the query type, response time guarantees can be offered in terms of the size of the compiled structure.

Numerous other examples, such as diagnosis and planning problems, can be formulated as graphical models and could benefit from compilation (Palacios, Bonet, Darwiche, & Geffner, 2005; Huang & Darwiche, 2005a). In diagnosis, compilation can facilitate fast detection of possible faults or explanations for some unusual behavior. Planning problems can also be formulated as graphical models, and a compilation would allow swift adjustments according to changes in the environment. Probabilistic models are one of the most used types of graphical models, and the basic query is to compute conditional probabilities of some variables given the evidence. A compact compilation of a probabilistic model would allow fast response to queries that incorporate evidence acquired in time. For example, two of the most important tasks for Bayesian networks are computing the probability of the evidence, and computing the maximum probable explanation (MPE). If some of the model variables become assigned (evidence), these tasks can be performed in time linear in the compilation size, which in practice is in many cases smaller than the upper-bound based on the treewidth or pathwidth of the graph. Formal verification is another example where compilation is heavily used to compare equivalence of circuit design, or to check the behavior of a circuit. *Binary Decision Diagram* (BDD) (Bryant, 1986) is arguably the most widely known and used compiled structure.

The contributions made in this paper to knowledge compilation in general and to decision diagrams in particular are the following:

1. We formally describe the AND/OR Multi-Valued Decision Diagram (AOMDD) and prove it to be a canonical representation for constraint networks, given a pseudo tree.

2. We extend the AOMDD to general weighted graphical models.

3. We give a compilation algorithm based on AND/OR search, that saves the trace of a memory intensive search and then reduces it in one bottom up pass.

4. We present the APPLY operator that combines two AOMDDs and show that its complexity is at most quadratic in the input, but never worse than exponential in the treewidth.

5. We give a scheduling order for building the AOMDD of a graphical model starting with the AOMDDs of its functions which is based on a Variable Elimination algorithm. This guarantees that the complexity is at most exponential in the *induced width* (treewidth) along the ordering.

6. We show how AOMDDs relate to various earlier and recent compilation frameworks, providing a unifying perspective for all these methods.

7. We introduce the *semantic treewidth*, which helps explain why compiled decision diagrams are often much smaller than the worst case bound.

8. We provide an experimental evaluation of the new data structure.

The structure of the paper is as follows. Section 2 provides preliminary definitions, a description of binary decision diagrams and the Bucket Elimination algorithm. Section 3 gives an overview of AND/OR search spaces. Section 4 introduces the AOMDD and discusses its properties. Section 5 describes a search-based algorithm for compiling the AOMDD. Section 6 presents a compilation algorithm based on a Bucket Elimination schedule and the APPLY operation. Section 7 proves that the AOMDD is a canonical representation for constraint networks given a pseudo tree, and Section 8 extends the AOMDD to weighted graphical models and proves their canonicity. Section 9 ties the canonicity to the new concept of semantic treewidth. Section 10 provides an experimental evaluation. Section 11 presents related work and Section 12 concludes the paper. All the proofs appear in an appendix.

## 2. Preliminaries

**Notations**    A reasoning problem is defined in terms of a set of variables taking values from finite domains and a set of functions defined over these variables. We denote variables or subsets of variables by uppercase letters (e.g., $X, Y, \ldots$) and values of variables by lower case letters (e.g., $x, y, \ldots$). Sets are usually denoted by bold letters, for example $\mathbf{X} = \{X_1, \ldots, X_n\}$ is a set of variables. An assignment $(X_1 = x_1, \ldots, X_n = x_n)$ can be abbreviated as $x = (\langle X_1, x_1 \rangle, \ldots, \langle X_n, x_n \rangle)$ or $x = (x_1, \ldots, x_n)$. For a subset of variables $\mathbf{Y}$, $D_{\mathbf{Y}}$ denotes the Cartesian product of the domains of variables in $\mathbf{Y}$. The projection of an assignment $x = (x_1, \ldots, x_n)$ over a subset $\mathbf{Y}$ is denoted by $x_{\mathbf{Y}}$ or $x[\mathbf{Y}]$. We will also denote by $Y = y$ (or $y$ for short) the assignment of values to variables in $\mathbf{Y}$ from their respective domains. We denote functions by letters $f$, $g$, $h$ etc., and the scope (set of arguments) of the function $f$ by $scope(f)$.

### 2.1 Graphical Models

DEFINITION **1 (graphical model)** *A graphical model* $\mathcal{M}$ *is a 4-tuple,* $\mathcal{M} = \langle \mathbf{X}, \mathbf{D}, \mathbf{F}, \otimes \rangle$, *where:*

1. $\mathbf{X} = \{X_1, \ldots, X_n\}$ *is a finite set of variables;*
2. $\mathbf{D} = \{D_1, \ldots, D_n\}$ *is the set of their respective finite domains of values;*
3. $\mathbf{F} = \{f_1, \ldots, f_r\}$ *is a set of positive real-valued discrete functions (i.e., their domains can be listed), each defined over a subset of variables* $\mathbf{S}_i \subseteq \mathbf{X}$*, called its scope, and denoted by* $scope(f_i)$*.*
4. $\otimes$ *is a combination operator*[1] *(e.g.,* $\otimes \in \{\prod, \sum, \bowtie\}$ *– product, sum, join), that can take as input two (or more) real-valued discrete functions, and produce another real-valued discrete function.*

*The graphical model represents the combination of all its functions:* $\otimes_{i=1}^r f_i$*.*

Several examples of graphical models appear later, for example: Figure 1 shows a constraint network and Figure 2 shows a belief network.

In order to define the equivalence of graphical models, it is useful to introduce the notion of *universal* graphical model that is defined by a single function.

DEFINITION **2 (universal equivalent graphical model)** *Given a graphical model* $\mathcal{M} = \langle \mathbf{X}, \mathbf{D}, \mathbf{F}_1, \otimes \rangle$ *the universal equivalent model of* $\mathcal{M}$ *is* $u(\mathcal{M}) = \langle \mathbf{X}, \mathbf{D}, \mathbf{F}_2 = \{\otimes_{f_i \in \mathbf{F}_1} f_i\}, \otimes \rangle$*.*

Two graphical models are **equivalent** if they represent the same function. Namely, if they have the same universal model.

DEFINITION **3 (weight of a full and a partial assignment)** *Given a graphical model* $\mathcal{M} = \langle \mathbf{X}, \mathbf{D}, \mathbf{F} \rangle$*, the weight of a full assignment* $x = (x_1, \ldots, x_n)$ *is defined by* $w(x) = \otimes_{f \in \mathbf{F}} f(x[scope(f)])$*. Given a subset of variables* $\mathbf{Y} \subseteq \mathbf{X}$*, the weight of a partial assignment* $y$ *is the combination of all the functions whose scopes are included in* $\mathbf{Y}$ *(denoted by* $\mathbf{F}_{\mathbf{Y}}$*) evaluated at the assigned values. Namely,* $w(y) = \otimes_{f \in \mathbf{F}_{\mathbf{Y}}} f(y[scope(f)])$*.*

**Consistency** For most graphical models, the range of the functions has a special zero value "0" that is *absorbing* relative to the combination operator (e.g., multiplication). Combining anything with "0" yields a "0". The "0" value expresses the notion of inconsistent assignments. It is a primary concept in constraint networks but can also be defined relative to other graphical models that have a "0" element.

DEFINITION **4 (consistent partial assignment, solution)** *Given a graphical model having a "0" element, a partial assignment is consistent if its cost is non-zero. A solution is a consistent assignment to all the variables.*

DEFINITION **5 (primal graph)** *The* primal graph *of a graphical model is an undirected graph that has variables as its vertices and an edge connects any two variables that appear in the scope of the same function.*

The primal graph captures the structure of the knowledge expressed by the graphical model. In particular, graph separation indicates independency of sets of variables given some assignments to other variables. All of the advanced algorithms for graphical models exploit the graphical structure, by using a heuristically good elimination order, a tree decomposition or some similar method. We will use the concept of pseudo tree, which resembles the tree rearrangements introduced by Freuder and Quinn (1985):

---

1. The combination operator can also be defined axiomatically (Shenoy, 1992).

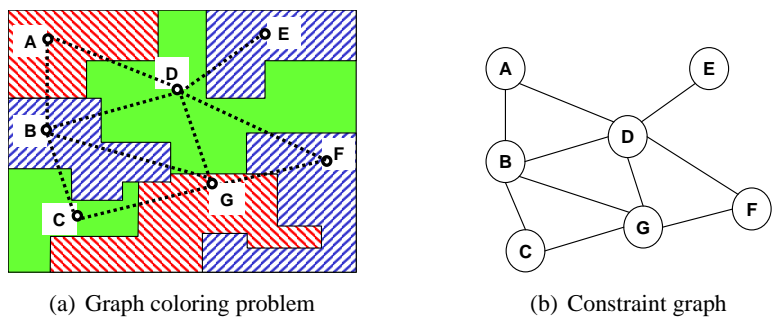

(a) Graph coloring problem        (b) Constraint graph

Figure 1: Constraint network

**DEFINITION 6 (pseudo tree)** *A* pseudo tree *of a graph* $G = (\mathbf{X}, E)$ *is a rooted tree* $\mathcal{T}$ *having the same set of nodes* $\mathbf{X}$, *such that every arc in* $E$ *is a backarc in* $\mathcal{T}$ *(A* path *in a rooted tree starts at the root and ends at one leaf. Two nodes can be connected by a* backarc *only if there exists a path that contains both).*

We use the common concepts and parameters from graph theory, that characterize the connectivity of the graph, and how close it is to a tree or to a chain. The induced width of a graphical model governs the complexity of solving it by Bucket Elimination (Dechter, 1999), and was also shown to bound the AND/OR search graph when memory is used to cache solved subproblems (Dechter & Mateescu, 2007).

**DEFINITION 7 (induced graph, induced width, treewidth, pathwidth)** *An* ordered *graph is a pair* $(G, d)$, *where* $G = (\{X_1, \ldots, X_n\}, E)$ *is an undirected graph, and* $d = (X_1, \ldots, X_n)$ *is an ordering of the nodes. The* width *of a node* in an ordered graph *is the number of neighbors that precede it in the ordering. The* width *of an ordering* $d$, *denoted* $w(d)$, *is the maximum width over all nodes. The* induced width *of an ordered graph,* $w^*(d)$, *is the width of the induced ordered graph obtained as follows: for each node, from last to first in* $d$, *its preceding neighbors are connected in a clique. The* induced width of a graph, $w^*$, *is the minimal induced width over all orderings. The induced width is also equal to the* treewidth *of a graph. The* pathwidth $pw^*$ *of a graph is the treewidth over the restricted class of orderings that correspond to chain decompositions.*

Various reasoning tasks, or queries can be defined over graphical models. Those can be defined formally using marginalization operators such as projection, summation and minimization. However, since our goal is to present a compilation of a graphical model which is independent of the queries that can be posed on it, we will discuss tasks in an informal manner only. For more information see the work of Kask, Dechter, Larrosa, and Dechter (2005).

Throughout the paper, we will use two examples of graphical models: constraint networks and belief networks. In the case of constraint networks, the functions can be understood as relations. In other words, the functions (also called constraints) can take only two values, $\{0, 1\}$, or $\{false, true\}$. A 0 value indicates that the corresponding assignment to the variables is inconsistent (not allowed), and a 1 value indicates consistency. Belief networks are an example of the more general case of graphical models (also called *weighted* graphical models). The functions in this case are conditional probability tables, so the values of a function are real numbers in the interval $[0, 1]$.

**Example 1** *Figure 1(a) shows a graph coloring problem that can be modeled by a constraint network. Given a map of regions, the problem is to color each region by one of the given colors {red, green, blue}, such that neighboring regions have different colors. The variables of the problems are the regions, and each one has the domain {red, green, blue}. The constraints are the relation "different" between neighboring regions. Figure 1(b) shows the constraint graph, and a solution (A=red, B=blue, C=green, D=green, E=blue, F=blue, G=red) is given in Figure 1(a). A more detailed example will be given later in Example 8.*

**Propositional Satisfiability** A special case of a CSP is *propositional satisfiability* (SAT). A formula $\varphi$ in *conjunctive normal form* (CNF) is a conjunction of *clauses* $\alpha_1, \ldots, \alpha_t$, where a clause is a disjunction of *literals* (propositions or their negations). For example, $\alpha = (P \vee \neg Q \vee \neg R)$ is a clause, where $P$, $Q$ and $R$ are propositions, and $P$, $\neg Q$ and $\neg R$ are literals. The SAT problem is to decide whether a given CNF theory has a *model*, *i.e.*, a truth-assignment to its propositions that does not violate any clause. Propositional satisfiability (SAT) can be defined as a CSP, where propositions correspond to variables, domains are $\{0, 1\}$, and constraints are represented by clauses, for example the clause $(\neg A \vee B)$ is a relation over its propositional variables that allows all tuples over $(A, B)$ except $(A = 1, B = 0)$.

**Cost Networks** An immediate extension of constraint networks are *cost networks* where the set of functions are real-valued cost functions, and the primary task is optimization. Also, GAI-nets (generalized additive independence, Fishburn, 1970) can be used to represent utility functions. An example of cost functions will appear in Figure 19.

DEFINITION **8 (cost network, combinatorial optimization)** *A* cost network *is a 4-tuple,* $\langle \mathbf{X}, \mathbf{D}, \mathbf{C}, \sum \rangle$*, where* $\mathbf{X}$ *is a set of variables* $\mathbf{X} = \{X_1, \ldots, X_n\}$*, associated with a set of discrete-valued domains,* $\mathbf{D} = \{D_1, \ldots, D_n\}$*, and a set of cost functions* $\mathbf{C} = \{C_1, \ldots, C_r\}$*. Each* $C_i$ *is a real-valued function defined on a subset of variables* $\mathbf{S}_i \subseteq \mathbf{X}$*. The combination operator, is* $\sum$*. The reasoning problem is to find a minimum cost solution.*

**Belief Networks** (Pearl, 1988) provide a formalism for reasoning about partial beliefs under conditions of uncertainty. They are defined by a directed acyclic graph over vertices representing random variables of interest (e.g., the temperature of a device, the gender of a patient, a feature of an object, the occurrence of an event). The arcs signify the existence of direct causal influences between linked variables quantified by conditional probabilities that are attached to each cluster of parents-child vertices in the network.

DEFINITION **9 (belief networks)** *A belief network (BN) is a graphical model* $\mathcal{P} = \langle \mathbf{X}, \mathbf{D}, \mathbf{P}_G, \prod \rangle$*, where* $\mathbf{X} = \{X_1, \ldots, X_n\}$ *is a set of variables over domains* $\mathbf{D} = \{D_1, \ldots, D_n\}$*. Given a directed acyclic graph G over* $\mathbf{X}$ *as nodes,* $P_G = \{P_1, \ldots, P_n\}$*, where* $P_i = \{P(X_i \,|\, pa\,(X_i))\}$ *are conditional probability tables (CPTs for short) associated with each* $X_i$*, where* $pa(X_i)$ *are the parents of* $X_i$ *in the acyclic graph G. A belief network represents a probability distribution over* $\mathbf{X}$*,* $P(x_1, \ldots, x_n) = \prod_{i=1}^{n} P(x_i | x_{pa(X_i)})$*. An evidence set* $e$ *is an instantiated subset of variables.*

When formulated as a graphical model, functions in $F$ denote conditional probability tables and the scopes of these functions are determined by the directed acyclic graph $G$: each function $f_i$ ranges over variable $X_i$ and its parents in $G$. The combination operator is product, $\otimes = \prod$. The primal graph of a belief network (viewed as an undirected model) is called a *moral graph*. It connects any two variables appearing in the same CPT.

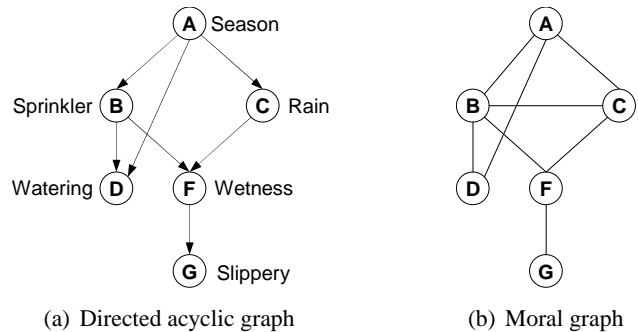

(a) Directed acyclic graph    (b) Moral graph

Figure 2: Belief network

**Example 2** *Figure 2(a) gives an example of a belief network over 6 variables, and Figure 2(b) shows its moral graph . The example expresses the causal relationship between variables "Season" (A), "The configuration of an automatic sprinkler system" (B), "The amount of rain expected" (C), "The amount of manual watering necessary" (D), "The wetness of the pavement" (F) and "Whether or not the pavement is slippery" (G). The belief network expresses the probability distribution $P(A, B, C, D, F, G) = P(A) \cdot P(B|A) \cdot P(C|A) \cdot P(D|B, A) \cdot P(F|C, B) \cdot P(G|F)$. Another example of a belief network and CPTs appears in Figure 9.*

The two most popular tasks for belief networks are defined below:

DEFINITION **10 (belief updating, most probable explanation (MPE))** *Given a belief network and evidence e, the* belief updating *task is to compute the posterior marginal probability of variable $X_i$, conditioned on the evidence. Namely,*

$$Bel(X_i = x_i) = P(X_i = x_i \mid e) = \alpha \sum_{\{(x_1, \ldots, x_{i-1}, x_{i+1}, \ldots, x_n) | E = e, X_i = x_i\}} \prod_{k=1}^{n} P(x_k, e | x_{pa_k}),$$

*where $\alpha$ is a normalization constant. The* most probable explanation (MPE) *task is to find a complete assignment which agrees with the evidence, and which has the highest probability among all such assignments. Namely, to find an assignment $(x_1^o, \ldots, x_n^o)$ such that*

$$P(x_1^o, \ldots, x_n^o) = max_{x_1, \ldots, x_n} \prod_{k=1}^{n} P(x_k, e | x_{pa_k}).$$

## 2.2 Binary Decision Diagrams Review

Decision diagrams are widely used in many areas of research to represent decision processes. In particular, they can be used to represent functions. Due to the fundamental importance of Boolean functions, a lot of effort has been dedicated to the study of *Binary Decision Diagrams* (BDDs), which are extensively used in software and hardware verification (Clarke et al., 1999; McMillan, 1993). The earliest work on BDDs is due to Lee (1959), who introduced the binary-decision *program*, that can be understood as a linear representation of a BDD (e.g., a depth first search ordering of the nodes), where each node is a branching instruction indicating the address of the next instruction for both the 0 and the 1 value of the test variable. Akers (1978) presented the actual graphical representation and further developed the BDD idea. However, it was Bryant (1986) that introduced what is now called the *Ordered Binary Decision Diagram* (OBDD). He restricted the order of variables along any path of the diagram, and presented algorithms (most importantly the *apply* procedure, that combines two OBDDs by an operation) that have time complexity at most quadratic in the sizes of the input diagrams. OBDDs are fundamental for applications with large binary functions, especially because in many practical cases they provide very compact representations.

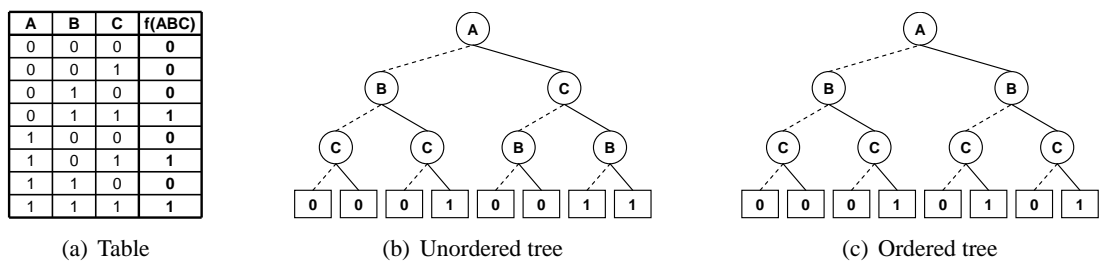

(a) Table  (b) Unordered tree  (c) Ordered tree

Figure 3: Boolean function representations

A BDD is a representation of a Boolean function. Given $\mathbf{B} = \{0, 1\}$, a Boolean function $f : \mathbf{B}^n \to \mathbf{B}$, has $n$ arguments, $X_1, \cdots, X_n$, which are Boolean variables, and takes Boolean values.

**Example 3** *Figure 3(a) shows a table representation of a Boolean function of three variables. This explicit representation is the most straightforward, but also the most costly due to its exponential requirements. The same function can also be represented by a binary tree, shown in Figure 3(b), that has the same exponential size in the number of variables. The internal round nodes represent the variables, the solid edges are the 1 (or high) value, and the dotted edges are the 0 (or low) value. The leaf square nodes show the value of the function for each assignment along a path. The tree shown in 3(b) is unordered, because variables do not appear in the same order along each path.*

In building an OBDD, the first condition is to have variables appear in the same order (A,B,C) along every path from root to leaves. Figure 3(c) shows an ordered binary tree for our function. Once an order is imposed, there are two reduction rules that transform a decision diagram into an equivalent one:

*(1) isomorphism:* merge nodes that have the same label and the same children.
*(2) redundancy:* eliminate nodes whose low and high edges point to the same node, and connect parent of removed node directly to child of removed node.

Applying the two reduction rules exhaustively yields a *reduced* OBDD, sometimes denoted rOBDD. We will just use OBDD and assume that it is completely reduced.

**Example 4** *Figure 4(a) shows the binary tree from Figure 3(c) after the isomorphic terminal nodes (leaves) have been merged. The highlighted nodes, labeled with C, are also isomorphic, and Figure 4(b) shows the result after they are merged. Now, the highlighted nodes labeled with C and B are redundant, and removing them gives the OBDD in Figure 4(c).*

## 2.3 Bucket Elimination Review

Bucket Elimination (**BE**) (Dechter, 1999) is a well known variable elimination algorithm for inference in graphical models. We will describe it using the terminology for constraint networks, but **BE** can also be applied to any graphical model. Consider a constraint network $\mathcal{R} = \langle \mathbf{X}, \mathbf{D}, \mathbf{C} \rangle$ and an ordering $d = (X_1, X_2, \ldots, X_n)$. The ordering $d$ dictates an elimination order for **BE**, from last to first. Each variable is associated with a bucket. Each constraint from $\mathbf{C}$ is placed in the bucket of its latest variable in $d$. Buckets are processed from $X_n$ to $X_1$ by eliminating the bucket variable (the constraints residing in the bucket are joined together, and the bucket variable is projected out) and placing the resulting constraint (also called *message*) in the bucket of its latest variable in $d$. After its execution, **BE** renders the network backtrack free, and a solution can be produced by assigning variables along $d$. **BE** can also produce the solutions count if marginalization is done by summation (rather than projection) over the functional representation of the constraints, and join is substituted by multiplication.

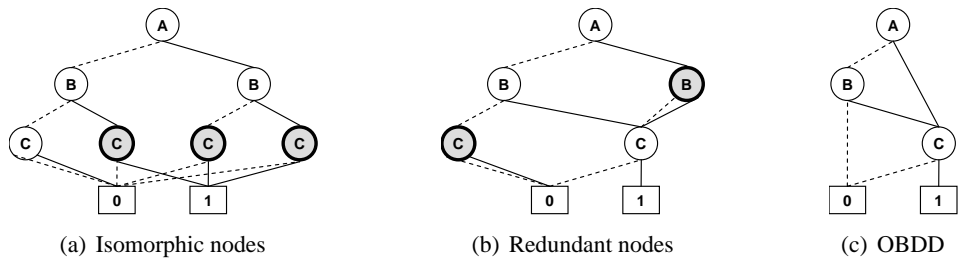

(a) Isomorphic nodes      (b) Redundant nodes      (c) OBDD

Figure 4: Reduction rules

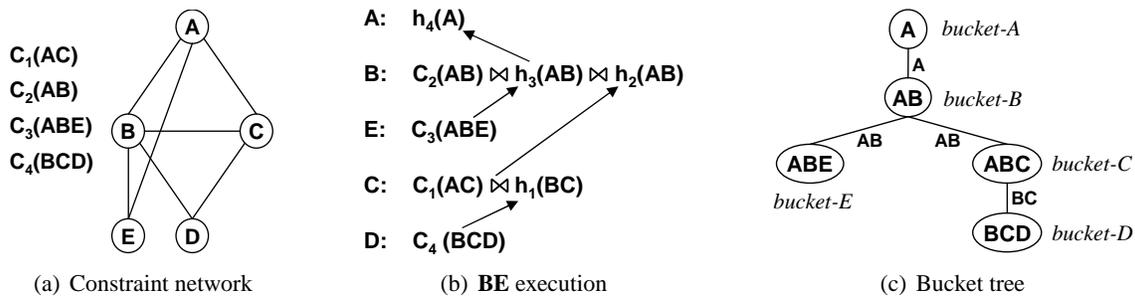

(a) Constraint network      (b) **BE** execution      (c) Bucket tree

Figure 5: Bucket Elimination

**BE** also constructs a bucket tree, by linking the bucket of each $X_i$ to the destination bucket of its message (called the parent bucket). A node in the bucket tree typically has a *bucket variable*, a *collection of constraints*, and a *scope* (the union of the scopes of its constraints). If the nodes of the bucket tree are replaced by their respective bucket variables, it is easy to see that we obtain a pseudo tree.

**Example 5** *Figure 5(a) shows a network with four constraints. Figure 5(b) shows the execution of Bucket Elimination along $d = (A, B, E, C, D)$. The buckets are processed from $D$ to $A$.[2] Figure 5(c) shows the bucket tree. The pseudo tree corresponding to the order $d$ is given in Fig. 6(a).*

---

2. The representation in Figure 5 reverses the top down bucket processing described in earlier papers (Dechter, 1999).

---

**Procedure** `GeneratePseudoTree`$(G, d)$

> **input** : graph $G = (\mathbf{X}, E)$; order $d = (X_1, \ldots, X_n)$
>
> **output** : Pseudo tree $\mathcal{T}$
>
> **1** Make $X_1$ the root of $\mathcal{T}$
>
> **2** Condition on $X_1$ (eliminate $X_1$ and its incident edges from $G$). Let $G_1, \ldots, G_p$ be the resulting connected components of $G$
>
> **3 for** $i = 1$ *to* $p$ **do**
>
> **4** $\quad$ $\mathcal{T}_i = $ `GeneratePseudoTree`$(G_i, d|_{G_i})$
>
> **5** $\quad$ Make root of $\mathcal{T}_i$ a child of $X_1$
>
> **6 return** $\mathcal{T}$

---

## 2.4 Orderings and Pseudo Trees

Given an ordering $d$, the structural information captured in the primal graph through the scopes of the functions $\mathbf{F} = \{f_1, \ldots, f_r\}$ can be used to create the unique pseudo tree that corresponds to $d$ (Mateescu & Dechter, 2005). This is precisely the bucket tree (or elimination tree), that is created by **BE** (when variables are processed in reverse $d$). The same pseudo tree can be created by conditioning on the primal graph, and processing variables in the order $d$, as described in Procedure `GeneratePseudoTree`. In the following, $d|_{G_i}$ is the restriction of the order $d$ to the nodes of the graph $G_i$.

## 3. Overview of AND/OR Search Space for Graphical Models

The AND/OR search space is a recently introduced (Dechter & Mateescu, 2004a, 2004b, 2007) unifying framework for advanced algorithmic schemes for graphical models. Its main virtue consists in exploiting independencies between variables during search, which can provide exponential speedups over traditional search methods oblivious to problem structure. Since AND/OR MDDs are based on AND/OR search spaces we need to provide a comprehensive overview for the sake of completeness.

### 3.1 AND/OR Search Trees

The AND/OR search tree is guided by a pseudo tree of the primal graph. The idea is to exploit the problem decomposition into independent subproblems during search. Assigning a value to a variable (also known as conditioning), is equivalent in graph terms to removing that variable (and its incident edges) from the primal graph. A partial assignment can therefore lead to the decomposition of the residual primal graph into independent components, each of which can be searched (or solved) separately. The pseudo tree captures precisely all these decompositions given an order of variable instantiation.

**Definition 11 (AND/OR search tree of a graphical model)** *Given a graphical model* $\mathcal{M} = \langle \mathbf{X}, \mathbf{D}, \mathbf{F} \rangle$, *its primal graph* $G$ *and a pseudo tree* $\mathcal{T}$ *of* $G$, *the associated AND/OR search tree has alternating levels of OR and AND nodes. The OR nodes are labeled* $X_i$ *and correspond to variables. The AND nodes are labeled* $\langle X_i, x_i \rangle$ *(or simply* $x_i$*) and correspond to value assignments. The structure of the AND/OR search tree is based on* $\mathcal{T}$. *The root is an OR node labeled with the root of* $\mathcal{T}$. *The children of an OR node* $X_i$ *are AND nodes labeled with assignments* $\langle X_i, x_i \rangle$ *that*

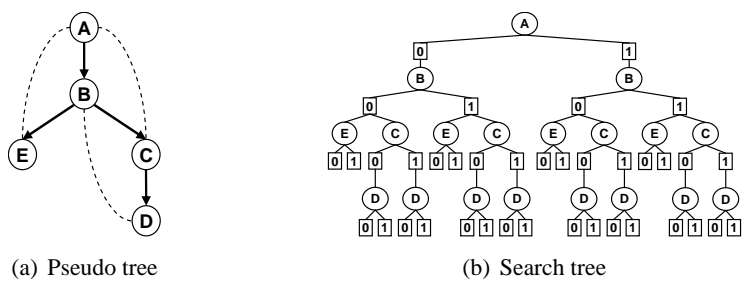

(a) Pseudo tree        (b) Search tree

Figure 6: AND/OR search tree

*are consistent with the assignments along the path from the root. The children of an AND node* $\langle X_i, x_i \rangle$ *are OR nodes labeled with the children of variable* $X_i$ *in the pseudo tree* $\mathcal{T}$.

**Example 6** *Figure 6 shows an example of an AND/OR search tree for the graphical model given in Figure 5(a), assuming all tuples are consistent, and variables are binary valued. When some tuples are inconsistent, some of the paths in the tree do not exist. Figure 6(a) gives the pseudo tree that guides the search, from top to bottom, as indicated by the arrows. The dotted arcs are backarcs from the primal graph. Figure 6(b) shows the AND/OR search tree, with the alternating levels of OR (circle) and AND (square) nodes, and having the structure indicated by the pseudo tree.*

The AND/OR search tree can be traversed by a depth first search algorithm, thus using linear space. It was already shown (Freuder & Quinn, 1985; Bayardo & Miranker, 1996; Darwiche, 2001; Dechter & Mateescu, 2004a, 2007) that:

**THEOREM 1** *Given a graphical model* $\mathcal{M}$ *over* $n$ *variables, and a pseudo tree* $\mathcal{T}$ *of depth* $m$, *the size of the AND/OR search tree based on* $\mathcal{T}$ *is* $O(n\,k^m)$, *where* $k$ *bounds the domains of variables. A graphical model of treewidth* $w^*$ *has a pseudo tree of depth at most* $w^* \log n$, *therefore it has an AND/OR search tree of size* $O(n\,k^{w^* \log n})$.

The AND/OR search tree expresses the set of all possible assignments to the problem variables (all solutions). The difference from the traditional OR search space is that a solution is no longer a path from root to a leaf, but rather a tree, defined as follows:

**DEFINITION 12 (solution tree)** *A solution tree of an AND/OR search tree contains the root node. For every OR node, it contains one of its child nodes and for each of its AND nodes it contains all its child nodes, and all its leaf nodes are consistent.*

## 3.2 AND/OR Search Graph

The AND/OR search tree may contain nodes that root identical subproblems. These nodes are said to be *unifiable*. When unifiable nodes are merged, the search space becomes a graph. Its size becomes smaller at the expense of using additional memory by the search algorithm. The depth first search algorithm can therefore be modified to cache previously computed results, and retrieve them when the same nodes are encountered again. The notion of unifiable nodes is defined formally next.

DEFINITION 13 (**minimal AND/OR graph, isomorphism**) *Two AND/OR search graphs $G$ and $G'$ are* isomorphic *if there exists a one to one mapping $\sigma$ from the vertices of $G$ to the vertices of $G'$ such that for any vertex $v$, if $\sigma(v) = v'$, then $v$ and $v'$ root identical subgraphs relative to $\sigma$. An AND/OR graph is called* minimal *if all its isomorphic subgraphs are merged. Isomorphic nodes (that root isomorphic subgraphs) are also said to be* unifiable.

It was shown by Dechter and Mateescu (2007) that:

THEOREM 2 *A graphical model $\mathcal{M}$ has a unique minimal AND/OR search graph relative to a pseudo-tree $\mathcal{T}$.*

The minimal AND/OR graph of a graphical model $\mathcal{G}$ relative to a pseudo tree $\mathcal{T}$ is denoted by $M_\mathcal{T}(G)$. Note that the definition of minimality used in the work of Dechter and Mateescu (2007) is based only on isomorphism reduction. We will extend it here by also including the elimination of redundant nodes. The previous theorem only shows that given an AND/OR graph, the merge operator has a fixed point, which is the minimal AND/OR graph. We will show in this paper that the AOMDD is a canonical representation, namely that any two equivalent graphical models can be represented by the same unique AOMDD given that they accept the same pseudo tree, and the AOMDD is minimal in terms of number of nodes.

Some unifiable nodes can be identified based on their *contexts*. We can define graph based contexts for both OR nodes and AND nodes, just by expressing the set of ancestor variables in $\mathcal{T}$ that completely determine a conditioned subproblem. However, it can be shown that using caching based on OR contexts makes caching based on AND contexts redundant and vice versa, so we will only use *OR caching*. Any value assignment to the context of $X$ separates the subproblem below $X$ from the rest of the network.

DEFINITION 14 (**OR context**) *Given a pseudo tree $\mathcal{T}$ of an AND/OR search space, $context(X) = [X_1 \ldots X_p]$ is the set of ancestors of $X$ in $\mathcal{T}$, ordered descendingly, that are connected in the primal graph to $X$ or to descendants of $X$.*

DEFINITION 15 (**context unifiable OR nodes**) *Given an AND/OR search graph, two OR nodes $n_1$ and $n_2$ are* context unifiable *if they have the same variable label $X$ and the assignments of their contexts is identical. Namely, if $\pi_1$ is the partial assignment of variables along the path to $n_1$, and $\pi_2$ is the partial assignment of variables along the path to $n_2$, then their restriction to the context of $X$ is the same: $\pi_1|_{context(X)} = \pi_2|_{context(X)}$.*

The depth first search algorithm that traverses the AND/OR search tree, can be modified to traverse a graph, if enough memory is available. We could allocate a cache table for each variable $X$, the scope of the table being $context(X)$. The size of the cache table for $X$ is therefore the product of the domains of variables in its context. For each variable $X$, and for each possible assignment to its context, the corresponding conditioned subproblem is solved only once and the computed value is saved in the cache table, and whenever the same context assignment is encountered again, the value of the subproblem is retrieved from the cache table. Such an algorithm traverses what is called the *context minimal AND/OR graph*.

DEFINITION 16 (**context minimal AND/OR graph**) *The* context minimal *AND/OR graph is obtained from the AND/OR search tree by merging all the context unifiable OR nodes.*

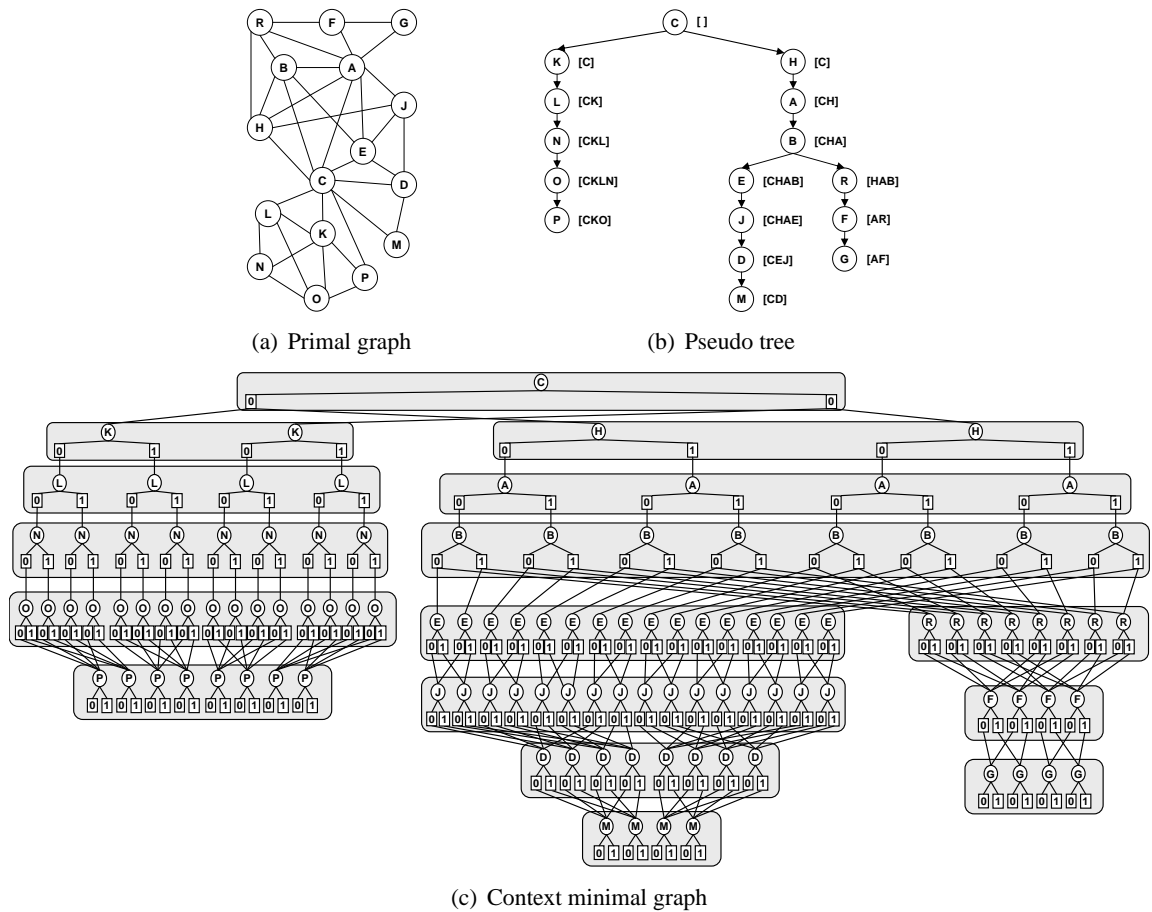

(a) Primal graph      (b) Pseudo tree

(c) Context minimal graph

Figure 7: AND/OR search graph

It was already shown (Bayardo & Miranker, 1996; Dechter & Mateescu, 2004a, 2007) that:

THEOREM 3 *Given a graphical model $\mathcal{M}$, its primal graph $G$ and a pseudo tree $\mathcal{T}$, the size of the context minimal AND/OR search graph based on $\mathcal{T}$, and therefore the size of its minimal AND/OR search graph, is $O(n\, k^{w^*_\mathcal{T}(G)})$, where $w^*_\mathcal{T}(G)$ is the induced width of $G$ over the depth first traversal of $\mathcal{T}$, and $k$ bounds the domain size.*

**Example 7** *Let's look at the impact of caching on the size of the search space by examining a larger example. Figure 7(a) shows a graphical model with binary variables and Figure 7(b) a pseudo tree that drives the AND/OR search. The context of each node is given in square brackets. The context minimal graph is given in Figure 7(c). Note that it is far smaller than the AND/OR search tree, which has $2^8 = 256$ AND nodes at the level of M alone (because M is at depth 8 in the pseudo tree). The shaded rectangles show the size of each cache table, equal to the number of OR nodes that appear in each one. A cache entry is useful whenever there are more than one incoming edges into the OR node. Incidentally, the caches that are not useful (namely OR nodes with only one incoming arc), are called* dead caches *(Darwiche, 2001), and can be determined based only on the pseudo* *tree inspection, therefore a cache table need not be allocated for them. The context minimal graph can also explain the execution of **BE** along the same pseudo tree (or, equivalently, along its depth first traversal order). The buckets are the shaded rectangles, and the processing is done bottom up. The number of possible assignments to each bucket equals the number of AND nodes that appear in it. The message scope is identical to the context of the bucket variable, and the message itself is identical to the corresponding cache table. For more details on the relationship between AND/OR search and **BE** see the work of Mateescu and Dechter (2005).*

### 3.3 Weighted AND/OR Graphs

In the previous subsections we described the structure of the AND/OR trees and graphs. In order to use them to solve a reasoning task, we need to define a way of using the input function values during the traversal of an AND/OR graph. This is realized by placing weights (or costs) on the OR-to-AND arcs, dictated by the function values. Only the functions that are relevant contribute to an OR-to-AND arc weight, and this is captured by the buckets relative to the pseudo tree:

**DEFINITION 17 (buckets relative to a pseudo tree)** *Given a graphical model $\mathcal{M} = \langle \mathbf{X}, \mathbf{D}, \mathbf{F}, \otimes \rangle$ and a pseudo tree $\mathcal{T}$, the* bucket *of $X_i$ relative to $\mathcal{T}$, denoted $B_{\mathcal{T}}(X_i)$, is the set of functions whose scopes contain $X_i$ and are included in $path_{\mathcal{T}}(X_i)$, which is the set of variables from the root to $X_i$ in $\mathcal{T}$. Namely,*

$$B_{\mathcal{T}}(X_i) = \{f \in \mathbf{F} | X_i \in scope(f), scope(f) \subseteq path_{\mathcal{T}}(X_i)\}.$$

A function belongs to the bucket of a variable $X_i$ iff its scope has just been fully instantiated when $X_i$ was assigned. Combining the values of all functions in the bucket, for the current assignment, gives the weight of the OR-to-AND arc:

**DEFINITION 18 (OR-to-AND weights)** *Given an AND/OR graph of a graphical model $\mathcal{M}$, the weight $w_{(n,m)}(X_i, x_i)$ of arc $(n, m)$ where $X_i$ labels $n$ and $x_i$ labels $m$, is the combination of all the functions in $B_{\mathcal{T}}(X_i)$ assigned by values along the current path to the AND node $m$, $\pi_m$. Formally, $w_{(n,m)}(X_i, x_i) = \otimes_{f \in B_{\mathcal{T}}(X_i)} f(asgn(\pi_m)[scope(f)]).$*

**DEFINITION 19 (weight of a solution tree)** *Given a weighted AND/OR graph of a graphical model $\mathcal{M}$, and given a solution tree $t$ having the OR-to-AND set of arcs $arcs(t)$, the weight of $t$ is defined by $w(t) = \otimes_{e \in arcs(t)} w(e).$*

**Example 8** *We start with the more straightforward case of constraint networks. Since functions only take values 0 or 1, and the combination is by product (join of relations), it follows that any OR-to-AND arc can only have a weight of 0 or 1. An example is given in Figure 8. Figure 8(a) shows a constraint graph, 8(b) a pseudo tree for it, and 8(c) the four relations that define the constraint problem. Figure 8(d) shows the AND/OR tree that can be traversed by a depth first search algorithm that only checks the consistency of the input functions (i.e., no constraint propagation is used). Similar to the OBDD representation, the OR-to-AND arcs with a weight of 0 are denoted by dotted lines, and the tree is not unfolded below them, since it will not contain any solution. The arcs with a weight of 1 are drawn with solid lines.*

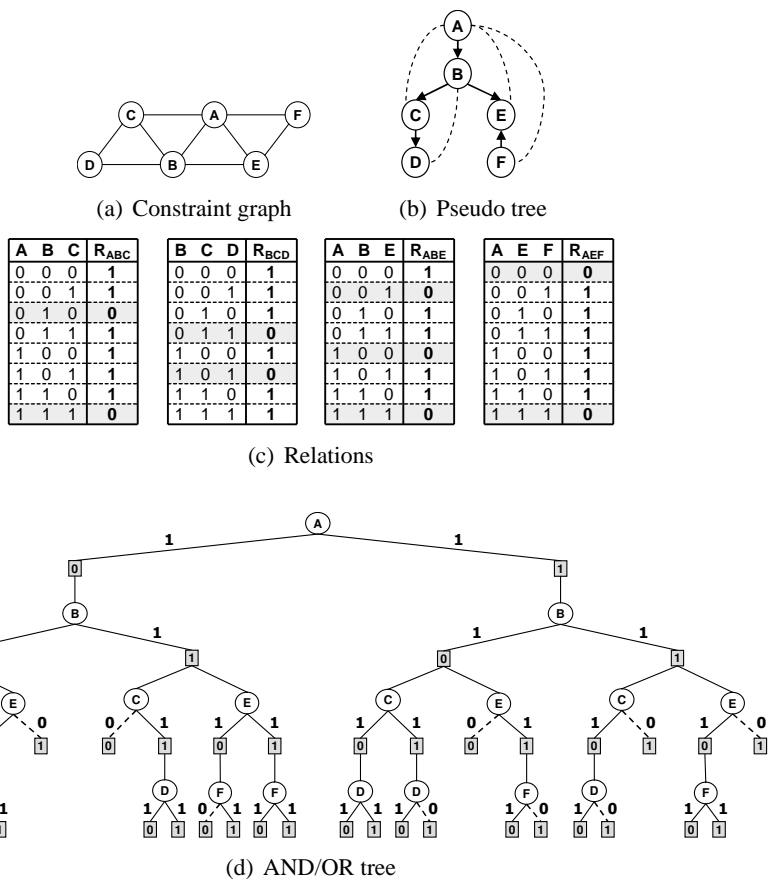

(a) Constraint graph   (b) Pseudo tree

(c) Relations

(d) AND/OR tree

Figure 8: AND/OR search tree for constraint networks

**Example 9** *Figure 9 shows a weighted AND/OR tree for a belief network. Figure 9(a) shows the directed acyclic graph, and the dotted arc BC added by moralization. Figure 9(b) shows the pseudo tree, and 9(c) shows the conditional probability tables. Figure 9(d) shows the weighted AND/OR tree.*

As we did for constraint networks, we can move from weighted AND/OR search trees to weighted AND/OR search graphs by merging unifiable nodes. In this case the arc labels should be also considered when determining unifiable subgraphs. This can yield context-minimal weighted AND/OR search graphs and minimal weighted AND/OR search graphs.

## 4. AND/OR Multi-Valued Decision Diagrams (AOMDDs)

In this section we begin describing the contributions of this paper. The *context minimal* AND/OR graph (Definition 16) offers an effective way of identifying some unifiable nodes during the execution of the search algorithm. Namely, context unifiable nodes are discovered based only on their paths from the root, without actually solving their corresponding subproblems. However, merging based on context is not complete, which means that there may still exist unifiable nodes in the search graph that do not have identical contexts. Moreover, some of the nodes in the context minimal AND/OR graph may be redundant, for example when the set of solutions rooted at variable $X_i$ is not dependant on the specific value assigned to $X_i$ (this situation is not detectable based on context). This is sometimes termed as "interchangeable values" or "symmetrical values". As overviewed earlier, Dechter and Mateescu (2007, 2004a) defined the complete *minimal AND/OR graph* which is an AND/OR graph whose unifiable nodes are all merged, and Dechter and Mateescu (2007) also proved the canonicity for non-weighted graphical models.

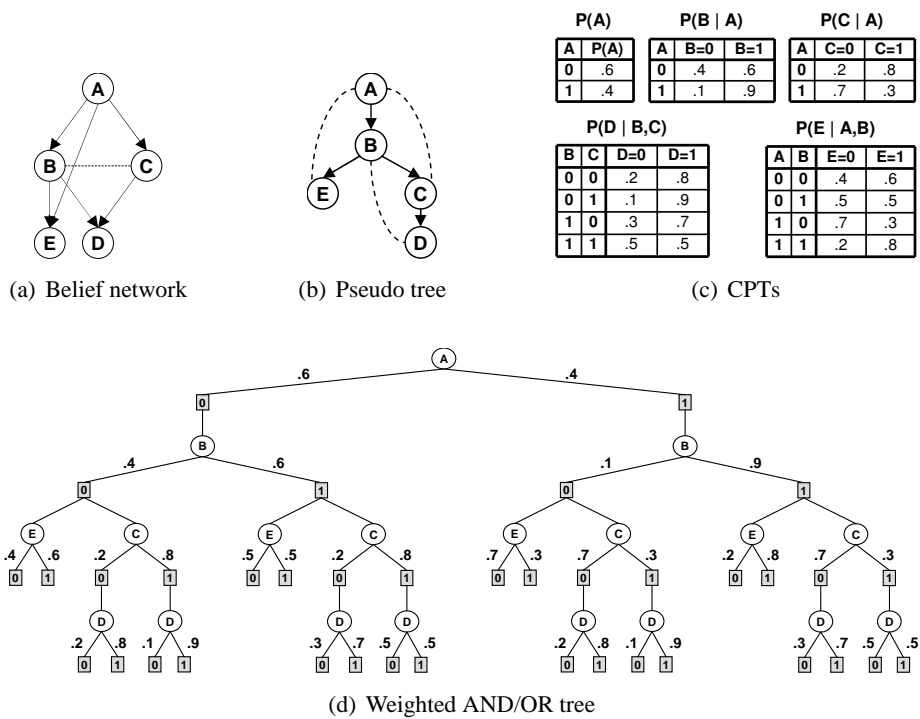

Figure 9: Weighted AND/OR search tree for belief networks

In this paper we propose to augment the minimal AND/OR search graph with removing redundant variables as is common in OBDD representation as well as adopt notational conventions common in this community. This yields a data structure that we call AND/OR BDD, that exploits decomposition by using AND nodes. We present the extension over multi-valued variables yielding AND/OR MDD or AOMDD and define them for general weighted graphical models. Subsequently we present two algorithms for compiling the canonical AOMDD of a graphical model: the first is search-based, and uses the memory intensive AND/OR graph search to generate the context minimal AND/OR graph, and then reduces it bottom up by applying reduction rules; the second is inference-based, and uses a Bucket Elimination schedule to combine the AOMDDs of initial functions by APPLY operations (similar to the *apply* for OBDDs). As we will show, both approaches have the same worst case complexity as the AND/OR graph search with context based caching, and also the same complexity as Bucket Elimination, namely time and space exponential in the treewidth of the problem, $O(n\,k^{w^*})$. The benefit of each of these generation schemes will be discussed.

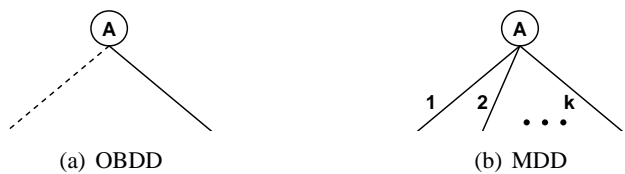

Figure 10: Decision diagram nodes (OR)

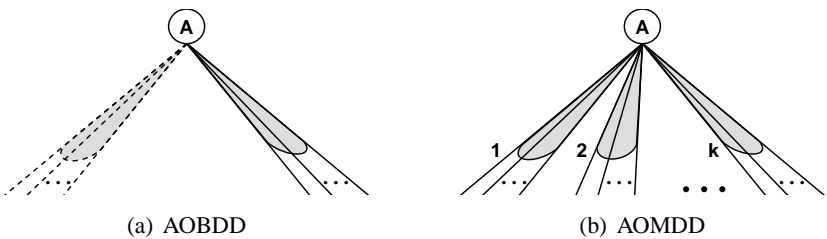

Figure 11: Decision diagram nodes (AND/OR)

## 4.1 From AND/OR Search Graphs to Decision Diagrams

An AND/OR search graph $\mathcal{G}$ of a graphical model $\mathcal{M} = \langle \mathbf{X}, \mathbf{D}, \mathbf{F}, \otimes \rangle$ represents the set of all possible assignments to the problem variables (all solutions and their costs). In this sense, $\mathcal{G}$ can be viewed as representing the function $f = \otimes_{f_i \in \mathbf{F}} f_i$ that defines the universal equivalent graphical model $u(\mathcal{M})$ (Definition 2). For each full assignment $x = (x_1, \ldots, x_n)$, if $x$ is a solution expressed by the tree $t_x$, then $f(x) = w(t_x) = \otimes_{e \in arcs(t_x)} w(e)$ (Definition 19); otherwise $f(x) = 0$ (the assignment is inconsistent). The solution tree $t_x$ of a consistent assignment $x$ can be read from $\mathcal{G}$ in linear time by following the assignments from the root. If $x$ is inconsistent, then a dead-end is encountered in $\mathcal{G}$ when attempting to read the solution tree $t_x$, and $f(x) = 0$. Therefore, $\mathcal{G}$ can be viewed as a decision diagram that determines the values of $f$ for every complete assignment $x$.

We will now see how we can process an AND/OR search graph by reduction rules similar to the case of OBDDs, in order to obtain a representation of minimal size. In the case of OBDDs, a node is labeled with a variable name, for example $A$, and the *low* (dotted line) and *high* (solid line) outgoing arcs capture the restriction of the function to the assignments $A = 0$ or $A = 1$. To determine the value of the function, one needs to follow either one or the other (but not both) of the outgoing arcs from $A$ (see Figure 10(a)). The straightforward extension of OBDDs to multi-valued variables (multi-valued decision diagrams, or MDDs) was presented by Srinivasan, Kam, Malik, and Brayton (1990), and the node structure that they use is given in Figure 10(b). Each outgoing arc is associated with one of the $k$ values of variable $A$.

In this paper we generalize the OBDD and MDD representations demonstrated in Figures 10(a) and 10(b) by allowing each outgoing arc to be an AND arc. An AND arc connects a node to a set of nodes, and captures the decomposition of the problem into independent components. The number of AND arcs emanating from a node is two in the case of AOBDDs (Figure 11(a)), or the domain size of the variable in the general case (Figure 11(b)). For a given node $A$, each of its $k$ AND arcs can connect it to possibly different number of nodes, depending on how the problem decomposes based on each particular assignment of $A$. The AND arcs are depicted by a shaded sector that connects the outgoing lines corresponding to the independent components.

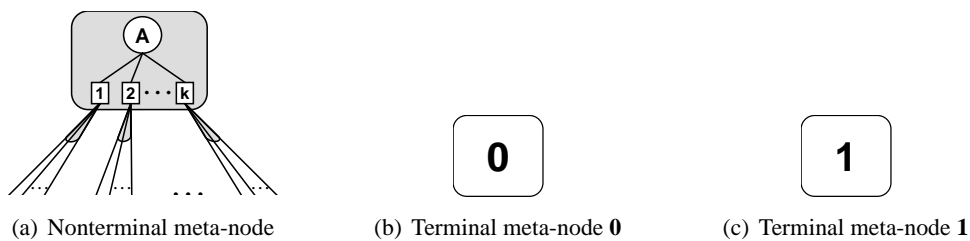

(a) Nonterminal meta-node    (b) Terminal meta-node **0**    (c) Terminal meta-node **1**

Figure 12: Meta-nodes

We define the AND/OR Decision Diagram representation based on AND/OR search graphs. We find that it is useful to maintain the semantics of Figure 11 especially when we need to express the redundancy of nodes, and therefore we introduce the *meta-node* data structure, which defines small portions of any AND/OR graph, based on an OR node and its AND children:

DEFINITION **20 (meta-node)** *A meta-node $u$ in an AND/OR search graph can be either: (1) a terminal node labeled with 0 or 1, or (2) a nonterminal node, that consists of an OR node labeled $X$ (therefore $var(u) = X$) and its $k$ AND children labeled $x_1, \ldots, x_k$ that correspond to the value assignments of $X$. Each AND node labeled $x_i$ stores a list of pointers to child meta-nodes, denoted by $u.children_i$. In the case of weighted graphical models, the AND node $x_i$ also stores the OR-to-AND arc weight $w(X, x_i)$.*

The rectangle in Figure 12(a) is a meta-node for variable $A$, that has a domain of size $k$. Note that this is very similar to Figure 11, with the small difference that the information about the value of $A$ that corresponds to each outgoing AND arc is now stored in the AND nodes of the meta-node. We are not showing the weights in that figure. A larger example of an AND/OR graph with meta-nodes appears later in Figure 16.

The terminal meta-nodes play the role of the terminal nodes in OBDDs. The terminal meta-node **0**, shown in Figure 12(b), indicates inconsistent assignments, while the terminal meta-node **1**, shown in figure 12(c) indicates consistent ones.

Any AND/OR search graph can now be viewed as a diagram of meta-nodes, simply by grouping OR nodes with their AND children, and adding the terminal meta-nodes appropriately.

Once we have defined the meta-nodes, it is easier to see when a variable is redundant with respect to the outcome of the function based on the current partial assignment. A variable is redundant if any of its assignments leads to the same set of solutions.

DEFINITION **21 (redundant meta-node)** *Given a weighted AND/OR search graph $\mathcal{G}$ represented with meta-nodes, a meta-node $u$ with $var(u) = X$ and $|D(X)| = k$ is redundant iff:*

*(a) $u.children_1 = \ldots = u.children_k$ and*

*(b) $w(X, x_1) = \ldots = w(X, x_k)$.*

An AND/OR graph $\mathcal{G}$, that contains a redundant meta-node $u$, can be transformed into an equivalent graph $\mathcal{G}'$ by replacing any incoming arc into $u$ with its common list of children $u.children_1$, absorbing the common weight $w(X, x_1)$ by combination into the weight of the parent meta-node corresponding to the incoming arc, and then removing $u$ and its outgoing arcs from $\mathcal{G}$. The value $X = x_1$ is picked here arbitrarily, because they are all isomorphic. If $u$ is the root of the

---

**Procedure RedundancyReduction**

    **input**   : AND/OR graph $\mathcal{G}$; redundant meta-node $u$, with $var(u) = X$; List of meta-node parents of $u$,
              denoted by $Parents(u)$.

    **output** : Reduced AND/OR graph $\mathcal{G}$ after the elimination of $u$.

**1**   **if** $Parents(u)$ *is empty* **then**

**2**      **return** *independent AND/OR graphs rooted by meta-nodes in $u.children_1$, and constant $w(X, x_1)$*

**3**   **forall** $v \in Parents(u)$ *(assume $var(v) == Y$)* **do**

**4**      **forall** $i \in \{1, \ldots, |D(Y)|\}$ **do**

**5**          **if** $u \in v.children_i$ **then**

**6**              $v.children_i \leftarrow v.children_i \setminus \{u\}$

**7**              $v.children_i \leftarrow v.children_i \cup u.children_1$

**8**              $w(Y, y_i) \leftarrow w(Y, y_i) \otimes w(X, x_1)$

**9**   remove $u$

**10**   **return** *reduced AND/OR graph $\mathcal{G}$*

---

**Procedure IsomorphismReduction**

    **input**   : AND/OR graph $\mathcal{G}$; isomorphic meta-nodes $u$ and $v$; List of meta-node parents of $u$, denoted by
              $Parents(u)$.

    **output** : Reduced AND/OR graph $\mathcal{G}$ after the merging of $u$ and $v$.

**1**   **forall** $p \in Parents(u)$ **do**

**2**      **if** $u \in p.children_i$ **then**

**3**          $p.children_i \leftarrow p.children_i \setminus \{u\}$

**4**          $p.children_i \leftarrow p.children_i \cup \{v\}$

**5**   remove $u$

**6**   **return** *reduced AND/OR graph $\mathcal{G}$*

---

graph, then the common weight $w(X, x_1)$ has to be stored separately as a constant. Procedure `RedundancyReduction` formalizes the redundancy elimination.

DEFINITION **22 (isomorphic meta-nodes)** *Given a weighted AND/OR search graph $\mathcal{G}$ represented with meta-nodes, two meta-nodes $u$ and $v$ having $var(u) = var(v) = X$ and $|D(X)| = k$ are* isomorphic *iff:*

    *(a) $u.children_i = v.children_i \;\forall i \in \{1, \ldots, k\}$ and*

    *(b) $w^u(X, x_i) = w^v(X, x_i) \;\forall i \in \{1, \ldots, k\}$, (where $w^u$, $w^v$ are the weights of $u$ and $v$).*

Procedure `IsomorphismReduction` formalizes the process of merging isomorphic meta-nodes. Naturally, the AND/OR graph obtained by merging isomorphic meta-nodes is equivalent to the original one. We can now define the AND/OR Multi-Valued Decision Diagram:

DEFINITION **23 (AOMDD)** *An AND/OR Multi-Valued Decision Diagram (AOMDD) is a weighted AND/OR search graph that is completely reduced by isomorphic merging and redundancy removal, namely:*

    *(1) it contains no isomorphic meta-nodes; and*

    *(2) it contains no redundant meta-nodes.*

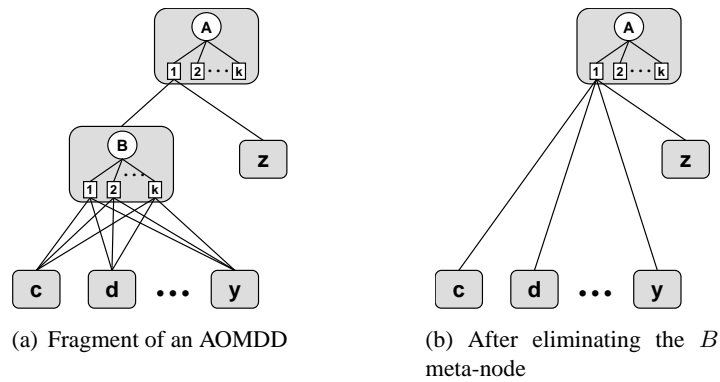

(a) Fragment of an AOMDD

(b) After eliminating the $B$ meta-node

Figure 13: Redundancy reduction

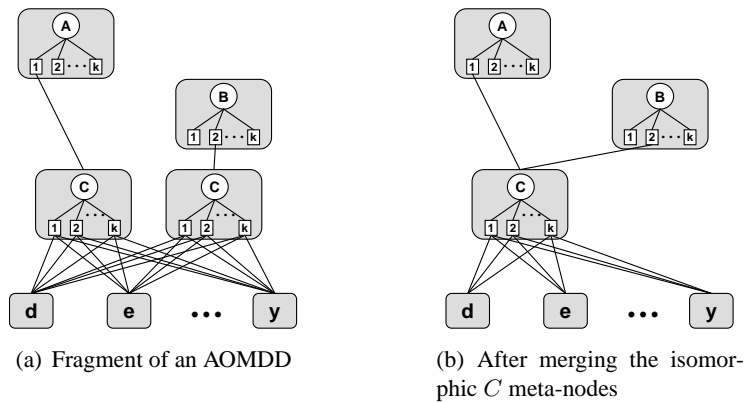

(a) Fragment of an AOMDD

(b) After merging the isomorphic $C$ meta-nodes

Figure 14: Isomorphism reduction

**Example 10** *Figure 13 shows an example of applying the redundancy reduction rule to a portion of an AOMDD. On the left side, in Figure 13(a), the meta-node of variable $B$ is redundant (we don't show the weights of the OR-to-AND arcs, to avoid cluttering the figure). Any of the values $\{1, \ldots, k\}$ of $B$ will lead to the same set of meta-nodes $\{c, d, \ldots, y\}$, which are coupled in an AND arc. Therefore, the meta-node of $B$ can be eliminated. The result is shown in Figure 13(b), where the meta-nodes $\{c, d, \ldots, y\}$ and $z$ are coupled in an AND arc outgoing from $A = 1$.*

*In Figure 14 we show an example of applying the isomorphism reduction rule. In this case, the meta-nodes labeled with $C$ in Figure 14(a) are isomorphic (again, we omit the weights). The result of merging them is shown in Figure 14(b).*

Examples of AOMDDs appear in Figures 16, 17 and 18. Note that if the weight on an OR-to-AND arc is zero, then the descendant is the terminal meta-node **0**. Namely, the current path is a dead-end, cannot be extended to a solution, and is therefore linked directly to **0**.

## 5. Using AND/OR Search to Generate AOMDDs

In Section 4.1 we described how we can transform an AND/OR graph into an AOMDD by applying reduction rules. In Section 5.1 we describe the explicit algorithm that takes as input a graphi-cal model, performs AND/OR search with context-based caching to obtain the context minimal AND/OR graph, and in Section 5.2 we give the procedure that applies the reduction rules bottom up to obtain the AOMDD.

### 5.1 Algorithm AND/OR-SEARCH-AOMDD

Algorithm 1, called AND/OR-SEARCH-AOMDD, compiles a graphical model into an AOMDD. A memory intensive (with context-based caching) AND/OR search is used to create the context minimal AND/OR graph (see Definition 16). The input to AND/OR-SEARCH-AOMDD is a graphical model $\mathcal{M}$ and a pseudo tree $\mathcal{T}$, that also defines the OR-context of each variable.

Each variable $X_i$ has an associated cache table, whose scope is the context of $X_i$ in $\mathcal{T}$. This ensures that the trace of the search is the context minimal AND/OR graph. A list denoted by $L^{X_i}$ (see line 35), is used for each variable $X_i$ to save pointers to meta-nodes labeled with $X_i$. These lists are used by the procedure that performs the bottom up reduction, per layers of the AND/OR graph (one layer contains all the nodes labeled with one given variable). The fringe of the search is maintained on a stack called OPEN. The current node (either OR or AND node) is denoted by n, its parent by p, and the current path by $\pi_n$. The children of the current node are denoted by $successors(\mathtt{n})$. For each node n, the Boolean attribute $consistent(\mathtt{n})$ indicates if the current path can be extended to a solution. This information is useful for pruning the search space.

The algorithm is based on two mutually recursive steps: **Forward** (beginning at line 5) and **Backtrack** (beginning at line 29), which call each other (or themselves) until the search terminates. In the forward phase, the AND/OR graph is expanded top down. The two types of nodes, AND and OR, are treated differently according to their semantics.

Before an OR node is expanded, the cache table of its variable is checked (line 8). If the entry is not null, a link is created to the already existing OR node that roots the graph equivalent to the current subproblem. Otherwise, the OR node is expanded by generating its AND descendants. The OR-to-AND weight (see Definition 18) is computed in line 13. Each value $x_i$ of $X_i$ is checked for consistency (line 14). The least expensive check is to verify that the OR-to-AND weight is non-zero. However, the deterministic (inconsistent) assignments in $\mathcal{M}$ can be extracted to form a constraint network. Any level of constraint propagation can be performed in this step (e.g., look ahead, arc consistency, path consistency, i-consistency etc.). The computational overhead can increase, in the hope of pruning the search space more aggressively. We should note that constraint propagation is not crucial for the algorithm, and the complexity guarantees are maintained even if only the simple weight check is performed. The consistent AND nodes are added to the list of successors of n (line 16), while the inconsistent ones are linked to the terminal **0** meta-node (line 19).

An AND node n labeled with $\langle X_i, x_i \rangle$ is expanded (line 20) based on the structure of the pseudo tree. If $X_i$ is a leaf in $\mathcal{T}$, then n is linked to the terminal **1** meta-node (line 22). Otherwise, an OR node is created for each child of $X_i$ in $\mathcal{T}$ (line 24).

The forward step continues as long as the current node is not a dead-end and still has unevaluated successors. The backtrack phase is triggered when a node has an empty set of successors (line 29). Note that, as each successor is processed, it is removed from the set of successors in line 42. When the backtrack reaches the root (line 32), the search is complete, the context minimal AND/OR graph is generated, and the Procedure BOTTOMUPREDUCTION is called.

When the backtrack step processes an OR node (line 31), it saves a pointer to it in cache, and also adds a pointer to the corresponding meta-node to the list $L^{X_i}$. The $consistent$ attribute of

---

**Algorithm 1**: AND/OR Search - AOMDD

---

    **input**    : $\mathcal{M} = \langle \mathbf{X}, \mathbf{D}, \mathbf{F} \rangle$; pseudo tree $\mathcal{T}$ rooted at $X_1$; parents $pa_i$ (OR-context) for every variable $X_i$.
    **output** : AOMDD of $\mathcal{M}$.

**1 forall** $X_i \in \mathbf{X}$ **do**
**2**     Initialize context-based cache table $Cache_{X_i}(pa_i)$ with **null** entries

**3** Create new $OR$ node $\mathtt{t}$, labeled with $X_i$; $consistent(\mathtt{t}) \leftarrow true$; push $\mathtt{t}$ on top of OPEN
**4 while** OPEN $\neq \phi$ **do**
**5**     $\mathtt{n} \leftarrow top(\mathtt{OPEN})$; remove $\mathtt{n}$ from OPEN                   // **Forward**
**6**     $successors(\mathtt{n}) \leftarrow \phi$
**7**     **if** $\mathtt{n}$ *is an OR node labeled with* $X_i$ **then**              // OR-expand
**8**        **if** $Cache_{X_i}(asgn(\pi_n)[pa_i]) \neq$ **null then**
**9**           Connect parent of $\mathtt{n}$ to $Cache_{X_i}(asgn(\pi_n)[pa_i])$    // Use the cached pointer
**10**        **else**
**11**           **forall** $x_i \in D_i$ **do**
**12**              Create new $AND$ node $\mathtt{t}$, labeled with $\langle X_i, x_i \rangle$
**13**              $w(X, x_i) \leftarrow \underset{f \in B_{\mathcal{T}}(X_i)}{\otimes} f(asgn(\pi_n)[pa_i])$
**14**              **if** $\langle X_i, x_i \rangle$ *is consistent* with $\pi_n$ **then**      // Constraint Propagation
**15**                 $consistent(\mathtt{t}) \leftarrow true$
**16**                 add $\mathtt{t}$ to $successors(\mathtt{n})$
**17**              **else**
**18**                 $consistent(\mathtt{t}) \leftarrow false$
**19**                 make terminal **0** the only child of $\mathtt{t}$

**20**     **if** $\mathtt{n}$ *is an AND node labeled with* $\langle X_i, x_i \rangle$ **then**          // AND-expand
**21**        **if** $children_{\mathcal{T}}(X_i) == \phi$ **then**
**22**           make terminal **1** the only child of $\mathtt{n}$
**23**        **else**
**24**           **forall** $Y \in children_{\mathcal{T}}(X_i)$ **do**
**25**              Create new $OR$ node $\mathtt{t}$, labeled with $Y$
**26**              $consistent(\mathtt{t}) \leftarrow false$
**27**              add $\mathtt{t}$ to $successors(\mathtt{n})$

**28**     Add $successors(\mathtt{n})$ to top of OPEN
**29**     **while** $successors(\mathtt{n}) == \phi$ **do**                      // **Backtrack**
**30**        let $\mathtt{p}$ be the parent of $\mathtt{n}$
**31**        **if** $\mathtt{n}$ *is an OR node labeled with* $X_i$ **then**
**32**           **if** $X_i == X_1$ **then**                    // Search is complete
**33**              Call BottomUpReduction procedure     // begin reduction to AOMDD
**34**           $Cache(asgn(\pi_n)[pa_i]) \leftarrow \mathtt{n}$               // Save in cache
**35**           Add meta-node of $\mathtt{n}$ to the list $L^{X_i}$
**36**           $consistent(\mathtt{p}) \leftarrow consistent(\mathtt{p}) \wedge consistent(\mathtt{n})$
**37**           **if** $consistent(\mathtt{p}) == false$ **then**        // Check if $\mathtt{p}$ is dead-end
**38**              remove $successors(\mathtt{p})$ from OPEN
**39**              $successors(\mathtt{p}) \leftarrow \phi$

**40**        **if** $\mathtt{n}$ *is an AND node labeled with* $\langle X_i, x_i \rangle$ **then**
**41**           $consistent(\mathtt{p}) \leftarrow consistent(\mathtt{p}) \vee consistent(\mathtt{n})$;

**42**        remove $\mathtt{n}$ from $successors(\mathtt{p})$
**43**        $\mathtt{n} \leftarrow \mathtt{p}$

---

**Procedure** `BottomUpReduction`

    **input**   : A graphical model $\mathcal{M} = \langle \mathbf{X}, \mathbf{D}, \mathbf{F} \rangle$; a pseudo tree $\mathcal{T}$ of the primal graph, rooted at $X_1$; Context minimal AND/OR graph, and lists $L^{X_i}$ of meta-nodes for each level $X_i$.

    **output** : AOMDD of $\mathcal{M}$.

**1** Let $d = \{X_1, \ldots, X_n\}$ be the depth first traversal ordering of $\mathcal{T}$

**2** **for** $i \leftarrow n$ **down to** 1 **do**

**3**      Let $H$ be a hash table, initially empty

**4**      **forall** *meta-nodes* n *in* $L^{X_i}$ **do**

**5**          **if** $H(X_i, n.children_1, \ldots, n.children_{k_i}, w^n(X_i, x_1), \ldots, w^n(X_{k_i}, x_{k_i}))$ *returns a meta-node* p **then**

**6**             merge n with p in the AND/OR graph

**7**          **else if** n *is redundant* **then**

**8**             eliminate n from the AND/OR graph

**9**             combine its weight with that of the parent

**10**         **else**

**11**             hash n into the table H:

**12**             $H(X_i, n.children_1, \ldots, n.children_{k_i}, w^n(X_i, x_1), \ldots, w^n(X_{k_i}, x_{k_i})) \leftarrow$ n

**13** **return** *reduced AND/OR graph*

---

the AND parent p is updated by conjunction with $consistent(\text{n})$. If the AND parent p becomes inconsistent, it is not necessary to check its remaining OR successors (line 38). When the backtrack step processes an AND node (line 40), the *consistent* attribute of the OR parent p is updated by disjunction with $consistent(\text{n})$.

The AND/OR search algorithm usually maintains a value for each node, corresponding to a task that is solved. We did not include values in our description because an AOMDD is just an equivalent representation of the original graphical model $\mathcal{M}$. Any task over $\mathcal{M}$ can be solved by a traversal of the AOMDD. It is however up to the user to include more information in the meta-nodes (e.g., number of solutions for a subproblem).

## 5.2 Reducing the Context Minimal AND/OR Graph to an AOMDD

Procedure `BottomUpReduction` processes the variables bottom up relative to the pseudo tree $\mathcal{T}$. We use the depth first traversal ordering of $\mathcal{T}$ (line 1), but any other bottom up ordering is as good. The outer *for* loop (starting at line 2) goes through each level of the context minimal AND/OR graph (where a level contains all the OR and AND nodes labeled with the same variable, in other words it contains all the meta-nodes of that variable). For efficiency, and to ensure the complexity guarantees that we will prove, a hash table, initially empty, is used for each level. The inner *for* loop (starting at line 4) goes through all the metanodes of a level, that are also saved (or pointers to them are saved) in the list $L^{X_i}$. For each new meta-node n in the list $L^{X_i}$, in line 5 the hash table $H$ is checked to verify if a node isomorphic with n already exists. If the hash table $H$ already contains a node p corresponding to the hash key $(X_i, n.children_1, \ldots, n.children_{k_i}, w^n(X_i, x_1), \ldots, w^n(X_{k_i}, x_{k_i}))$, then p and n are isomorphic and should be merged. Otherwise, if the new meta-node n is redundant, then it is eliminated from the AND/OR graph. If none of the previous two conditions is met, then the new meta-node n is hashed into table $H$.

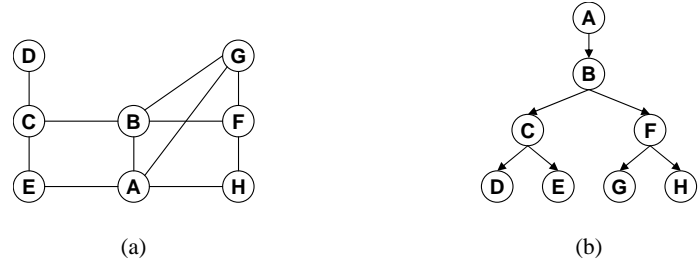

(a)                    (b)

Figure 15: (a) Constraint graph for $C = \{C_1, \ldots, C_9\}$, where $C_1 = F \vee H$, $C_2 = A \vee \neg H$, $C_3 = A \oplus B \oplus G$, $C_4 = F \vee G$, $C_5 = B \vee F$, $C_6 = A \vee E$, $C_7 = C \vee E$, $C_8 = C \oplus D$, $C_9 = B \vee C$; (b) Pseudo tree (bucket tree) for ordering $d = (A, B, C, D, E, F, G, H)$

**Proposition 1** *The output of Procedure* `BottomUpReduction` *is the AOMDD of* $\mathcal{M}$ *along the pseudo tree* $\mathcal{T}$, *namely the resulting AND/OR graph is completely reduced.*

Note that we explicated Procedure `BottomUpReduction` separately only for clarity. In practice, it can actually be included in Algorithm AND/OR-SEARCH-AOMDD, and the reduction rules can be applied whenever the search backtracks. We can maintain a hash table for each variable, during the AND/OR search, to store pointers to meta-nodes. When the search backtracks out of an OR node, it can already check the redundancy of that meta-node, and also look up in the hash table to check for isomorphism. Therefore, the reduction of the AND/OR graph can be done during the AND/OR search, and the output will be the AOMDD of $\mathcal{M}$.

From Theorem 3 and Proposition 1 we can conclude:

**THEOREM 4** *Given a graphical model* $\mathcal{M}$ *and a pseudo tree* $\mathcal{T}$ *of its primal graph* $G$, *the AOMDD of* $\mathcal{M}$ *corresponding to* $\mathcal{T}$ *has size bounded by* $O(n\ k^{w^*_{\mathcal{T}}(G)})$ *and it can be computed by Algorithm* AND/OR-SEARCH-AOMDD *in time* $O(n\ k^{w^*_{\mathcal{T}}(G)})$, *where* $w^*_{\mathcal{T}}(G)$ *is the induced width of* $G$ *over the depth first traversal of* $\mathcal{T}$, *and* $k$ *bounds the domain size.*

## 6. Using Bucket Elimination to Generate AOMDDs

In this section we propose to use a Bucket Elimination (**BE**) type algorithm to guide the compilation of a graphical model into an AOMDD. The idea is to express the graphical model functions as AOMDDs, and then combine them with APPLY operations based on a **BE** schedule. The APPLY is very similar to that from OBDDs (Bryant, 1986), but it is adapted to AND/OR search graphs. It takes as input two functions represented as AOMDDs based on the same pseudo tree, and outputs the combination of initial functions, also represented as an AOMDD based on the same pseudo tree. We will describe it in detail in Section 6.2.

We will start with an example based on constraint networks. This is easier to understand because the weights on the arcs are all 1 or 0, and therefore are depicted in the figures by solid and dashed lines, respectively.

**Example 11** *Consider the network defined by* $\mathbf{X} = \{A, B, \ldots, H\}$, $D_A = \ldots = D_H = \{0, 1\}$ *and the constraints (where* $\oplus$ *denotes XOR):* $C_1 = F \vee H$, $C_2 = A \vee \neg H$, $C_3 = A \oplus B \oplus G$, $C_4 = F \vee G$,

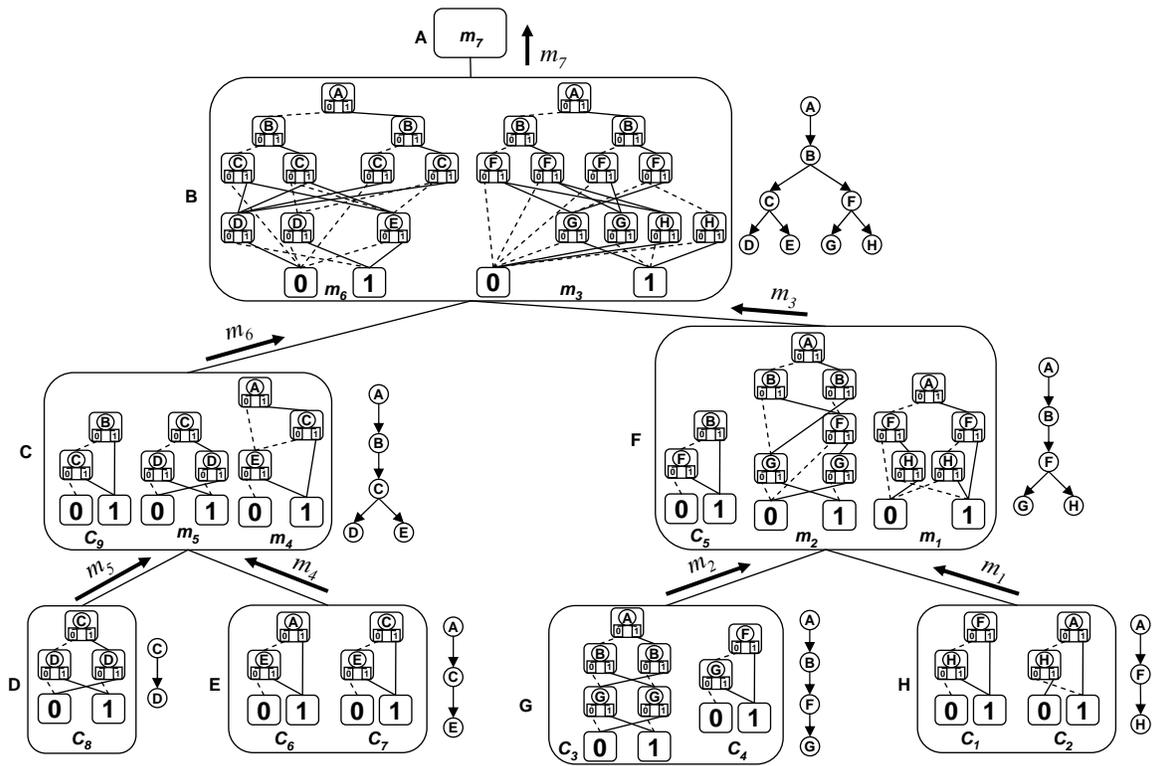

Figure 16: Execution of **BE** with AOMDDs

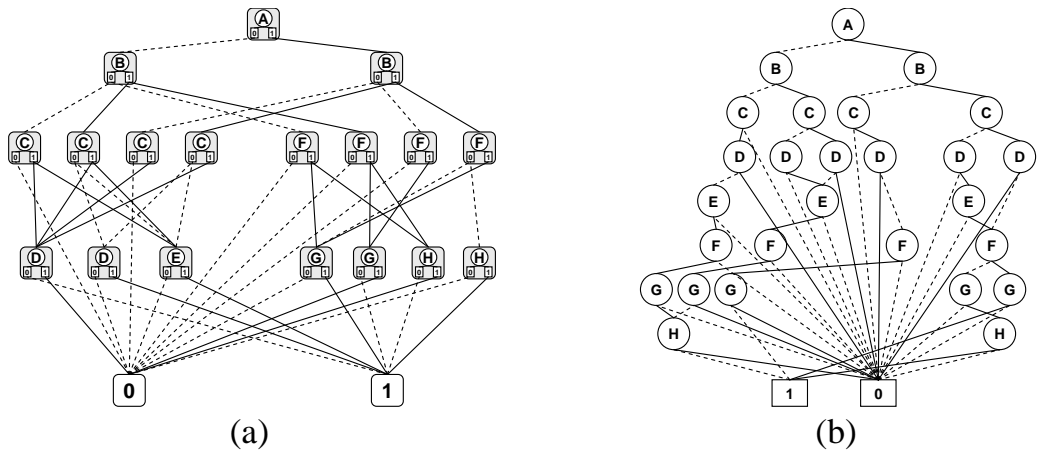

Figure 17: (a) The final AOMDD; (b) The OBDD corresponding to $d$

$C_5 = B \vee F$, $C_6 = A \vee E$, $C_7 = C \vee E$, $C_8 = C \oplus D$, $C_9 = B \vee C$. *The constraint graph is shown in Figure 15(a). Consider the ordering $d = (A, B, C, D, E, F, G, H)$. The pseudo tree (or bucket tree) induced by $d$ is given in Fig. 15(b). Figure 16 shows the execution of **BE** with AOMDDs along ordering $d$. Initially, the constraints $C_1$ through $C_9$ are represented as AOMDDs and placed in the bucket of their latest variable in $d$. The scope of any original constraint always appears on a*

---

**Algorithm 2**: BE-AOMDD

> **input** : Graphical model $\mathcal{M} = \langle \mathbf{X}, \mathbf{D}, \mathbf{F} \rangle$, where $\mathbf{X} = \{X_1, \ldots, X_n\}$, $\mathbf{F} = \{f_1, \ldots, f_r\}$ ; order
>          $d = (X_1, \ldots, X_n)$
>
> **output** : AOMDD representing $\otimes_{i \in \mathbf{F}} f_i$
>
> **1**   $\mathcal{T} = \texttt{GeneratePseudoTree}(G, d);$
> **2**   **for** $i \leftarrow 1$ **to** $r$ **do**                    // place functions in buckets
> **3**      place $\mathcal{G}_{f_i}^{aomdd}$ in the bucket of its latest variable in $d$
>
> **4**   **for** $i \leftarrow n$ **down to** $1$ **do**                      // process buckets
> **5**      $message(X_i) \leftarrow \mathcal{G}_{\mathbf{1}}^{aomdd}$            // initialize with AOMDD of **1**;
> **6**      **while** $bucket(X_i) \neq \phi$ **do**          // combine AOMDDs in bucket of $X_i$
> **7**          pick $\mathcal{G}_f^{aomdd}$ from $bucket(X_i);$
> **8**          $bucket(X_i) \leftarrow bucket(X_i) \setminus \{\mathcal{G}_f^{aomdd}\};$
> **9**          $message(X_i) \leftarrow \textsc{apply}(message(X_i), \mathcal{G}_f^{aomdd})$
> **10**     add $message(X_i)$ to the bucket of the parent of $X_i$ in $\mathcal{T}$
> **11**   **return** $message(X_1)$

---

*path from root to a leaf in the pseudo tree. Therefore, each* original *constraint is represented by an AOMDD based on a chain (i.e., there is no branching into independent components at any point). The chain is just the scope of the constraint, ordered according to d. For bi-valued variables, the original constraints are represented by OBDDs, for multiple-valued variables they are MDDs. Note that we depict meta-nodes: one OR node and its two AND children, that appear inside each gray node. The dotted edge corresponds to the 0 value (the* low *edge in OBDDs), the solid edge to the 1 value (the* high *edge). We have some redundancy in our notation, keeping both AND value nodes and arc-types (dotted arcs from "0" and solid arcs from "1").*

The **BE** *scheduling is used to process the buckets in reverse order of d. A bucket is processed by* joining *all the AOMDDs inside it, using the* APPLY *operator. However, the step of elimination of the bucket variable is omitted because we want to generate the full AOMDD. In our example, the messages* $m_1 = C_1 \bowtie C_2$ *and* $m_2 = C_3 \bowtie C_4$ *are still based on chains, therefore they are OBDDs. Note that they contain the variables H and G, which have not been eliminated. However, the message* $m_3 = C_5 \bowtie m_1 \bowtie m_2$ *is not an OBDD anymore. We can see that it follows the structure of the pseudo tree, where F has two children, G and H. Some of the nodes corresponding to F have two outgoing edges for value 1.*

The processing continues in the same manner. The final output of the algorithm, which coincides with $m_7$, is shown in Figure 17(a). The OBDD based on the same ordering d is shown in Fig. 17(b). Notice that the AOMDD has 18 nonterminal nodes and 47 edges, while the OBDD has 27 nonterminal nodes and 54 edges.

## 6.1 Algorithm BE-AOMDD

Algorithm 2, called BE-AOMDD, creates the AOMDD of a graphical model by using a **BE** schedule for APPLY operations. Given an order $d$ of the variables, first a pseudo tree is created based on the primal graph. Each initial function $f_i$ is then represented as an AOMDD, denoted by $\mathcal{G}_{f_i}^{aomdd}$, and placed in its bucket. To obtain the AOMDD of a function, the scope of the function is ordered according to $d$, a search tree (based on a chain) that represents $f_i$ is generated, and then reduced by Procedure BottomUpReduction. The algorithm proceeds exactly like **BE**, with the only difference that the combination of functions is realized by the APPLY algorithm, and variables are not eliminated but carried over to the destination bucket. The messages between buckets are initialized with the dummy AOMDD of **1**, denoted by $\mathcal{G}_1^{aomdd}$, which is neutral for combination.

In order to create the compilation of a graphical model based on AND/OR graphs, it is necessary to traverse the AND/OR graph top down and bottom up. This is similar to the inward and outward message passing in a tree decomposition. Note that BE-AOMDD describes the bottom up traversal explicitly, while the top down phase is actually performed by the APPLY operation. When two AOMDDs are combined, after the top chain portion of their pseudo tree is processed, the remaining independent branches are attached only if they participate in the newly restricted set of solutions. This amounts to an exchange of information between the independent branches, which is equivalent to the top down phase.

## 6.2 The AOMDD APPLY Operation

We will now describe how to combine two AOMDDs. The APPLY operator takes as input two AOMDDs representing functions $f_1$ and $f_2$ and returns an AOMDD representing $f_1 \otimes f_2$.

In OBDDs the *apply* operator combines two input diagrams based on the same variable ordering. Likewise, in order to combine two AOMDDs we assume that their pseudo trees are *identical*. This condition is satisfied by any two AOMDDs in the same bucket of BE-AOMDD. However, we present here a version of APPLY that is more general, by relaxing the previous condition from *identical* to *compatible* pseudo trees. Namely, there should be a pseudo tree in which both can be embedded. In general, a pseudo tree induces a strict partial order between the variables where a parent node always precedes its child nodes.

DEFINITION 24 (compatible pseudo trees) *A strict partial order $d_1 = (\mathbf{X}, <_1)$ over a set $\mathbf{X}$ is* consistent *with a strict partial order $d_2 = (\mathbf{Y}, <_2)$ over a set $\mathbf{Y}$, if for all $x_1, x_2 \in \mathbf{X} \cap \mathbf{Y}$, if $x_1 <_2 x_2$ then $x_1 <_1 x_2$. Two partial orders $d_1$ and $d_2$ are* compatible *iff there exists a partial order $d$ that is consistent with both. Two pseudo trees are* compatible *iff the partial orders induced via the parent-child relationship, are compatible.*

For simplicity, we focus on a more restricted notion of compatibility, which is sufficient when using a **BE** like schedule for the APPLY operator to combine the input AOMDDs (as described in Section 6). The APPLY algorithm that we will present can be extended to the more general notion of compatibility.

DEFINITION 25 (strictly compatible pseudo trees) *A pseudo tree $\mathcal{T}_1$ having the set of nodes $\mathbf{X}_1$ can be* embedded *in a pseudo tree $\mathcal{T}$ having the set of nodes $\mathbf{X}$ if $\mathbf{X}_1 \subseteq \mathbf{X}$ and $\mathcal{T}_1$ can be obtained from $\mathcal{T}$ by deleting each node in $\mathbf{X} \setminus \mathbf{X}_1$ and connecting its parent to each of its descendents. Two pseudo trees $\mathcal{T}_1$ and $\mathcal{T}_2$ are* strictly compatible *if there exists $\mathcal{T}$ such that both $\mathcal{T}_1$ and $\mathcal{T}_2$ can be embedded in $\mathcal{T}$.*

Algorithm APPLY (algorithm 3) takes as input one node from $\mathcal{G}_f^{aomdd}$ and a list of nodes from $\mathcal{G}_g^{aomdd}$. Initially, the node from $\mathcal{G}_f^{aomdd}$ is its root node, and the list of nodes from $\mathcal{G}_g^{aomdd}$ is in fact also made of just one node, which is its root. We will sometimes identify an AOMDD by its root node. The pseudo trees $\mathcal{T}_f$ and $\mathcal{T}_g$ are strictly compatible, having a target pseudo tree $\mathcal{T}$.

The list of nodes from $\mathcal{G}_g^{aomdd}$ always has a special property: there is no node in it that can be the ancestor in $\mathcal{T}$ of another (we refer to the variable of the meta-node). Therefore, the list $z_1, \ldots, z_m$

---

**Algorithm 3**: APPLY($v_1; z_1, \ldots, z_m$)

> **input** : AOMDDs $\mathcal{G}_f^{aomdd}$ with nodes $v_i$ and $\mathcal{G}_g^{aomdd}$ with nodes $z_j$, based on *strictly compatible* pseudo
> trees $\mathcal{T}_f$, $\mathcal{T}_g$ that can be embedded in $\mathcal{T}$.
> $var(v_1)$ is an ancestor of all $var(z_1), \ldots, var(z_m)$ in $\mathcal{T}$.
> $var(z_i)$ and $var(z_j)$ are not in ancestor-descendant relation in $\mathcal{T}, \forall i \neq j$.
>
> **output** : $v_1 \otimes (z_1 \wedge \ldots \wedge z_m)$, based on $\mathcal{T}$.

**1** **if** $H_1(v_1, z_1, \ldots, z_m) \neq null$ **then return** $H_1(v_1, z_1, \ldots, z_m)$;       // is in cache
**2** **if** *(any of* $v_1, z_1, \ldots, z_m$ *is* 0*)* **then return** 0
**3** **if** *(*$v_1 = 1$*)* **then return** 1
**4** **if** *(*$m = 0$*)* **then return** $v_1$       // nothing to combine
**5** create new nonterminal meta-node $u$
**6** $var(u) \leftarrow var(v_1)$ (call it $X_i$, with domain $D_i = \{x_1, \ldots, x_{k_i}\}$ )
**7** **for** $j \leftarrow 1$ **to** $k_i$ **do**
**8**     $u.children_j \leftarrow \phi$       // children of the j-th AND node of u
**9**     $w^u(X_i, x_j) \leftarrow w^{v_1}(X_i, x_j)$       // assign weight from v1
**10**     **if** *(* *(*$m = 1$*) and (*$var(v_1) = var(z_1) = X_i$*) )* **then**
**11**        $tempChildren \leftarrow z_1.children_j$
**12**        $w^u(X_i, x_j) \leftarrow w^{v_1}(X_i, x_j) \otimes w^{z_1}(X_i, x_j)$       // combine input weights
**13**     **else**
**14**        $tempChildren \leftarrow \{z_1, \ldots, z_m\}$
**15**     group nodes from $v_1.children_j \cup tempChildren$ in several $\{v^1; z^1, \ldots, z^r\}$
**16**     **for** *each* $\{v^1; z^1, \ldots, z^r\}$ **do**
**17**        $y \leftarrow$ APPLY($v^1; z^1, \ldots, z^r$)
**18**        **if** *(*$y = 0$*)* **then**
**19**           $u.children_j \leftarrow \mathbf{0}$; break
**20**        **else**
**21**           $u.children_j \leftarrow u.children_j \cup \{y\}$
**22**     **if** *(*$u.children_1 = \ldots = u.children_{k_i}$*) and (*$w^u(X_i, x_1) = \ldots = w^u(X_i, x_{k_i})$*)* **then**
**23**        promote $w^u(X_i, x_1)$ to parent
**24**        **return** $u.children_1$       // redundancy
**25**     **if** *(*$H_2(X_i, u.children_1, \ldots, u.children_{k_i}, w^u(X_i, x_1), \ldots, w^u(X_{k_i}, x_{k_i})) \neq null$*)* **then**
**26**        **return** $H_2(X_i, u.children_1, \ldots, u.children_{k_i}, w^u(X_i, x_1), \ldots, w^u(X_{k_i}, x_{k_i}))$
       // isomorphism
**27** Let $H_1(v_1, z_1, \ldots, z_m) = u$       // add $u$ to $H_1$
**28** Let $H_2(X_i, u.children_1, \ldots, u.children_{k_i}, w^u(X_i, x_1), \ldots, w^u(X_{k_i}, x_{k_i})) = u$   // add $u$ to $H_2$
**29** **return** $u$

---

from $g$ expresses a decomposition with respect to $\mathcal{T}$, so all those nodes appear on different branches. We will employ the usual techniques from OBDDs to make the operation efficient. First, if one of the arguments is **0**, then we can safely return **0**. Second, a hash table $H_1$ is used to store the nodes that have already been processed, based on the nodes $(v_1, z_1, \ldots, z_r)$. Therefore, we never need to make multiple recursive calls on the same arguments. Third, a hash table $H_2$ is used to detect isomorphic nodes. This is typically split in separate tables for each variable. If at the end of the recursion, before returning a value, we discover that a meta-node with the same variable, the same children and the same weights has already been created, then we don't need to store it and we simply return the existing node. And fourth, if at the end of the recursion we discover that we created a redundant node (all the children are the same and all the weights are the same), then we don't store it, and return instead one of its identical lists of children, and promote the common weight.

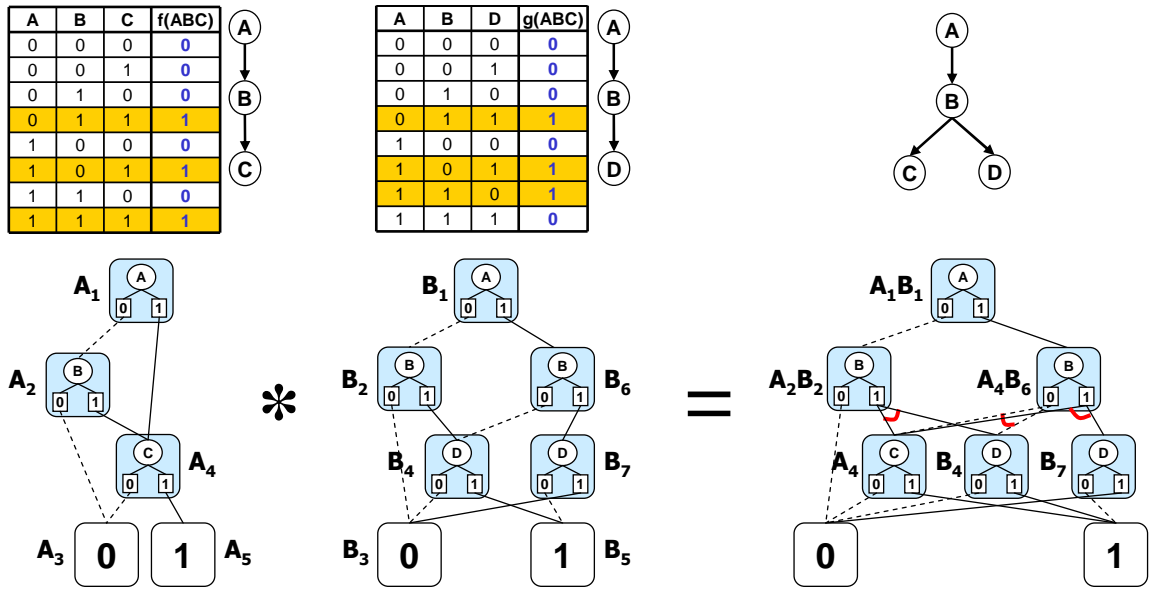

Figure 18: Example of APPLY operation

Note that $v_1$ is always an ancestor of all $z_1, \ldots, z_m$ in $\mathcal{T}$. We consider a variable in $\mathcal{T}$ to be an ancestor of itself. A few self explaining checks are performed in lines 1-4. Line 2 is specific for multiplication, and needs to be changed for other combination operations. The algorithm creates a new meta-node $u$, whose variable is $var(v_1) = X_i$ – recall that $var(v_1)$ is highest (closest to root) in $\mathcal{T}$ among $v_1, z_1, \ldots, z_m$. Then, for each possible value of $X_i$, line 7, it starts building its list of children.

One of the important steps happens in line 15. There are two lists of meta-nodes, one from each original AOMDD $f$ and $g$, and we will refer only to their variables, as they appear in $\mathcal{T}$. Each of these lists has the important property mentioned above, that its nodes are not ancestors of each other. The union of the two lists is grouped into maximal sets of nodes, such that the highest node in each set is an ancestor of all the others. It follows that the root node in each set belongs to one of the original AOMDD, say $v^1$ is from $f$, and the others, say $z^1, \ldots, z^r$ are from $g$. As an example, suppose $\mathcal{T}$ is the pseudo tree from Fig. 15(b), and the two lists are $\{C, G, H\}$ from $f$ and $\{E, F\}$ from $g$. The grouping from line 15 will create $\{C; E\}$ and $\{F; G, H\}$. Sometimes, it may be the case that a newly created group contains only one node. This means there is nothing more to join in recursive calls, so the algorithm will return, via line 4, the single node. From there on, only one of the input AOMDDs is traversed, and this is important for the complexity of APPLY, discussed below.

**Example 12** *Figure 18 shows the result of combining two Boolean functions by an AND operation (or product). The input functions $f$ and $g$ are represented by AOMDDs based on chain pseudo trees, while the results is based on the pseudo tree that expresses the decomposition after variables $A$ and $B$ are instantiated. The APPLY operator performs a depth first traversal of the two input AOMDDs, and generates the resulting AOMDD based on the output pseudo tree. Similar to the case of OBDDs, a function or an AOMDD can be identified by its root meta-node. In this example the input meta-nodes have labels $(A_1, A_2, B_1, B_2, \text{ etc.})$. The output meta-node labeled by $A_2B_2$ is* *the root of a diagram that represents the function obtained by combining the functions rooted by $A_2$ and $B_2$.*

### 6.3 Complexity of APPLY and BE-AOMDD

We now provide a characterization of the complexity of APPLY, based on different criteria. The following propositions are inspired by the results that govern OBDD *apply* complexity, but are adapted for pseudo tree orderings.

An AOMDD along a pseudo tree can be regarded as a union of regular MDDs, each restricted to a full path from root to a leaf in the pseudo tree. Let $\pi_{\mathcal{T}}$ be such a path in $\mathcal{T}$. Based on the definition of strictly compatible pseudo trees, $\pi_{\mathcal{T}}$ has corresponding paths $\pi_{\mathcal{T}_f}$ in $\mathcal{T}_f$ and $\pi_{\mathcal{T}_g}$ in $\mathcal{T}_g$. The MDDs from $f$ and $g$ corresponding to $\pi_{\mathcal{T}_f}$ and $\pi_{\mathcal{T}_g}$ can be combined using the regular MDD *apply*. This process can be repeated for every path $\pi_{\mathcal{T}}$. The resulting MDDs, one for each path in $\mathcal{T}$ need to be synchronized on their common parts (on the intersection of the paths). The algorithm we proposed does all this processing at once, in a depth first search traversal over the inputs. Based on our construction, we can give a first characterization of the complexity of AOMDD APPLY as being governed by the complexity of MDD *apply*.

**Proposition 2** *Let $\pi_1, \ldots, \pi_l$ be the set of paths in $\mathcal{T}$ enumerated from left to right and let $\mathcal{G}_f^i$ and $\mathcal{G}_g^i$ be the MDDs restricted to path $\pi_i$, then the size of the output of AOMDD* apply *is bounded by $\sum_i |\mathcal{G}_f^i| \cdot |\mathcal{G}_g^i| \leq n \cdot max_i |\mathcal{G}_f^i| \cdot |\mathcal{G}_g^i|$. The time complexity is also bounded by $\sum_i |\mathcal{G}_f^i| \cdot |\mathcal{G}_g^i| \leq n \cdot max_i |\mathcal{G}_f^i| \cdot |\mathcal{G}_g^i|$.*

A second characterization of the complexity can be given, similar to the MDD case, in terms of total number of nodes of the inputs:

**Proposition 3** *Given two AOMDDs $\mathcal{G}_f^{aomdd}$ and $\mathcal{G}_g^{aomdd}$ based on strictly compatible pseudo trees, the size of the output of* APPLY *is at most $O(|\mathcal{G}_f^{aomdd}| \cdot |\mathcal{G}_g^{aomdd}|)$.*

We can further detail the previous proposition as follows. Given AOMDDs $\mathcal{G}_f^{aomdd}$ and $\mathcal{G}_g^{aomdd}$, based on compatible pseudo trees $\mathcal{T}_f$ and $\mathcal{T}_g$ and the common pseudo tree $\mathcal{T}$, we define the *intersection pseudo tree* $\mathcal{T}_{f \cap g}$ as being obtained from $\mathcal{T}$ by the following two steps: (1) mark all the subtrees whose nodes belong to either $\mathcal{T}_f$ or $\mathcal{T}_g$ but not to both (the leaves of each subtree should be leaves in $\mathcal{T}$); (2) remove the subtrees marked in step (1) from $\mathcal{T}$. Steps (1) and (2) are applied just once (that is, not recursively). The part of AOMDD $\mathcal{G}_f^{aomdd}$ corresponding to the variables in $\mathcal{T}_{f \cap g}$ is denoted by $\mathcal{G}_f^{f \cap g}$, and similarly for $\mathcal{G}_g^{aomdd}$ it is denoted by $\mathcal{G}_g^{f \cap g}$.

**Proposition 4** *The time complexity of* APPLY *and the size of the output are $O(|\mathcal{G}_f^{f \cap g}| \cdot |\mathcal{G}_g^{f \cap g}| + |\mathcal{G}_f^{aomdd}| + |\mathcal{G}_g^{aomdd}|)$.*

We now turn to the complexity of the BE-AOMDD algorithm. Each bucket has an associated bucket pseudo tree. The top chain of the bucket pseudo tree for variable $X_i$ contains all and only the variables in $context(X_i)$. For any other variables that appear in the bucket pseudo tree, their associated buckets have already been processed. The original functions that belong to the bucket of $X_i$ have their scope included in $context(X_i)$, and therefore their associated AOMDDs are based on chains. Any other functions that appear in bucket of $X_i$ are messages received from independent branches below. Therefore, any two functions in the bucket of $X_i$ only share variables in the $context(X_i)$, which forms the top chain of the bucket pseudo tree. We can therefore characterize the complexity of APPLY in terms of treewidth, or context size of a bucket variable.

**Proposition 5** *Given two AOMDDs in the same bucket of* BE-AOMDD*, the time and space complexity of the* APPLY *between them is at most exponential in the context size of the bucket variable (namely the number of the variables in the top chain of the bucket pseudo tree).*

We can now bound the complexity of BE-AOMDD and the output size:

**THEOREM 5** *The space complexity of* BE-AOMDD *and the size of the output AOMDD are* $O(n\, k^{w^*})$*, where* $n$ *is the number of variables,* $k$ *is the maximum domain size and* $w^*$ *is the treewidth of the bucket tree. The time complexity is bounded by* $O(r\, k^{w^*})$*, where* $r$ *is the number of initial functions.*

## 7. AOMDDs Are Canonical Representations

It is well known that OBDDs are canonical representations of Boolean functions given an ordering of the variables (Bryant, 1986), namely a strict ordering of any CNF specification of the same Boolean function will yield an identical OBDD, and this property extends to MDDs (Srinivasan et al., 1990). The linear ordering of the variables defines a *chain* pseudo tree that captures the structure of the OBDD or MDD. In the case of AOBDDs and AOMDDs, the canonicity is with respect to a pseudo tree, transitioning from total orders (that correspond to a linear ordering) to partial orders (that correspond to a pseudo tree ordering). On the one hand we gain the ability to have a more compact compiled structure, but on the other hand canonicity is no longer with respect to all equivalent graphical models, but only relative to those graphical models that are consistent with the pseudo tree that is used. Specifically, if we start from a strict ordering we can generate a chain AOMDD that will be canonical relative to all equivalent graphical models. If however we want to exploit additional decomposition we can use a partial ordering captured by a pseudo-tree and create a more compact AOMDD. This AOMDD however is canonical relative to those equivalent graphical models that can accept the same pseudo tree that guided the AOMDD. In general, AOMDD can be viewed as a more flexible framework for compilation that allows both partial and total orderings. Canonicity is restricted to a subset of graphical models whose primal graph agrees with the partial order but it is relevant to a larger set of orderings which are consistent with the pseudo-tree.

In the following subsection we discuss the canonicity of AOMDD for constraint networks. The case of general weighted graphical models is discussed in Section 8.

### 7.1 AOMDDs for Constraint Networks Are Canonical Representations

The case of constraint networks is more straightforward, because the weights on the OR-to-AND arcs can only be 0 or 1. We will show that any equivalent constraint networks, that admit the same pseudo tree $\mathcal{T}$, have the same AOMDD based on $\mathcal{T}$. We start with a proposition that will help prove the main theorem.

**Proposition 6** *Let $f$ be a function, not always zero, defined by a constraint network over* **X***. Given a partition* $\{\mathbf{X}^1, \ldots, \mathbf{X}^m\}$ *of the set of variables* **X** *(namely,* $\mathbf{X}^i \cap \mathbf{X}^j = \phi, \forall\, i \neq j$*, and* $\mathbf{X} =$

$\cup_{i=1}^m \mathbf{X}^i$), if $f = f_1 \otimes \ldots \otimes f_m$ and $f = g_1 \otimes \ldots \otimes g_m$, such that $scope(f_i) = scope(g_i) = \mathbf{X}^i$ for all $i \in \{1, \ldots, m\}$, then $f_i = g_i$ for all $i \in \{1, \ldots, m\}$. Namely, if $f$ can be decomposed over the given partition, then the decomposition is unique.

We are now ready to show that AOMDDs for constraint networks are canonical representations given a pseudo tree.

THEOREM 6 (AOMDDs are canonical for a given pseudo tree) *Given a constraint network, and a pseudo tree $\mathcal{T}$ of its constraint graph, there is a unique (up to isomorphism) AOMDD that represents it, and it has the minimal number of meta-nodes.*

A constraint network is defined by its relations (or functions). There exist equivalent constraint networks that are defined by different sets of functions, even having different scope signatures. However, equivalent constraint networks define the same function, and we can ask if the AOMDD of different equivalent constraint networks is the same. The following corollary can be derived immediately from Theorem 6.

**Corollary 1** *Two equivalent constraint networks that admit the same pseudo tree $\mathcal{T}$ have the same AOMDD based on $\mathcal{T}$.*

## 8. Canonical AOMDDs for Weighted Graphical Models

Theorem 6 ensures that the AOMDD is canonical for constraint networks, namely for functions that can only take the values 0 or 1. The proof relied on the fact that the OR-to-AND weights can only be 0 or 1, and on Proposition 6 that ensured the unique decomposition of a function defined by a constraint network.

In this section we turn to general weighted graphical models. We can first observe that Proposition 6 is no longer valid for general functions. This is because the valid solutions (having strictly positive weight) can have their weight decomposed in more than one way into a product of positive weights.

Therefore we raise the issue of recognizing nodes that root AND/OR graphs that represent the same universal function, even though the graphical representation is different. We will see that the AOMDD for a weighted graphical model is not unique under the current definitions, but we can slightly modify them to obtain canonicity again. We have to note that canonicity of AOMDDs for weighted graphical models (e.g., belief networks) is far less crucial than in the case of OBDDs that are used in formal verification. Even more than that, sometimes it may be useful not to eliminate the redundant nodes, in order to maintain a simpler semantics of the AND/OR graph that represents the model.

The loss of canonicity of AOMDD for weighted graphical models can happen because of the weights on the OR-to-AND arcs, and we suggest a possible way of re-enforcing it if a more compact and canonical representation is needed.

**Example 13** *Figure 19 shows a weighted graphical model, defined by two (cost) functions, $f(M, A, B)$ and $g(M, B, C)$. Assuming the order (M,A,B,C), Figure 20 shows the AND/OR search tree on the left. The arcs are labeled with function values, and the leaves show the value of the corresponding full assignment (which is the product of numbers on the arcs of the path). We can*

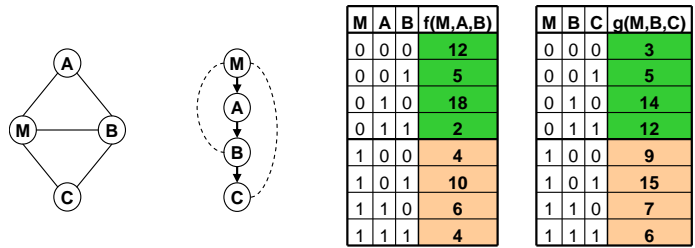

Figure 19: Weighted graphical model

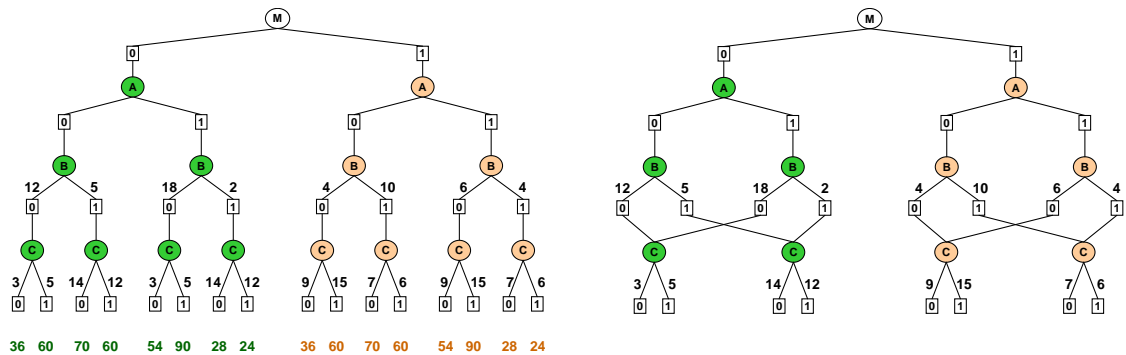

Figure 20: AND/OR search tree and context minimal graph

*see that either value of M (0 or 1) gives rise to the same function (because the leaves in the two subtrees have the same values). However, the two subtrees can not be identified as representing the same function by the usual reduction rules. The right part of the figure shows the context minimal graph, which has a compact representation of each subtree, but does not share any of their parts.*

What we would like in this case is to have a method of recognizing that the left and right subtrees corresponding to $M = 0$ and $M = 1$ represent the same function. We can do this by normalizing the values in each level, and processing bottom up. In Figure 21 left, the values on the OR-to-AND arcs have been normalized, for each OR variable, and the normalization constant was promoted up to the OR value. In Figure 21 right, the normalization constants are promoted upwards again by multiplication. This process does not change the value of each full assignment, and therefore produces equivalent graphs.

We can see already that some of the nodes labeled by C can now be merged, producing the graph in Figure 22 on the left. Continuing the same process we obtain the AOMDD for the weighted graph, shown in Figure 22 on the right.

We can define the AOMDD of a weighted graphical model as follows:

**DEFINITION 26 (AOMDD of weighted graphical model)** *The AOMDD of a weighted graphical model is an AND/OR graph, with meta-nodes, such that: (1) for each meta-node, its weights sum to 1; (2) the root meta-node has a constant associated with it; (3) it is completely reduced, namely it has no isomorphic meta-nodes, and no redundant meta-nodes.*

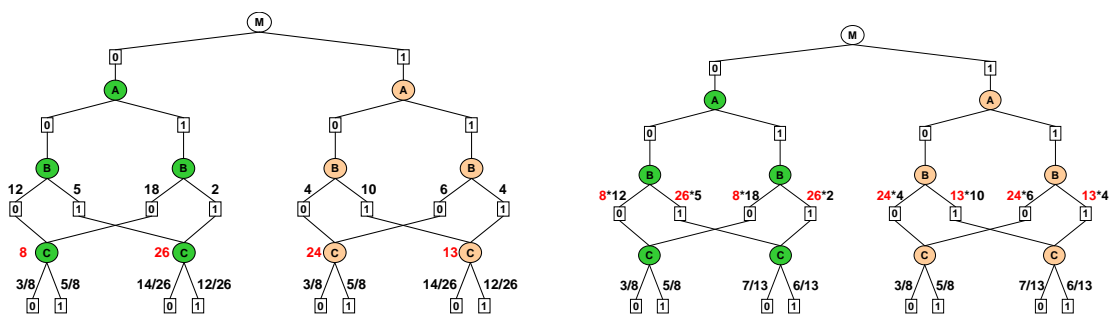

Figure 21: Normalizing values bottom up

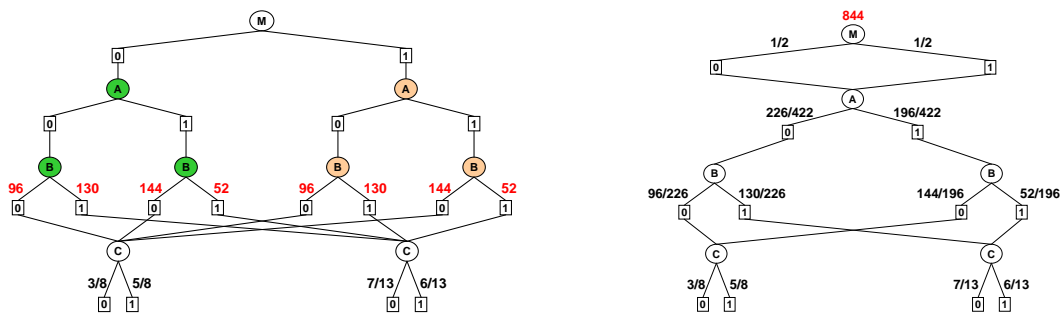

Figure 22: AOMDD for the weighted graph

The procedure of transforming a weighted AND/OR graph into an AOMDD is very similar to Procedure BOTTOMUPREDUCTION from Section 5. The only difference is that when a new layer is processed, first the meta-node weights are normalized and promoted to the parent, and then the procedure continues as usual with the reduction rules.

THEOREM 7 *Given two equivalent weighted graphical models that accept a common pseudo tree $\mathcal{T}$, normalizing arc values together with exhaustive application of reduction rules yields the same AND/OR graph, which is the AOMDD based on $\mathcal{T}$.*

**Finite Precision Arithmetic**   The implementation of the algorithm described in this section may prove to be challenging on machines that used finite precision arithmetic. Since the weights are real-valued, the repeated normalization may lead to precision errors. One possible approach, which we also used in our experiments, is to define some $\varepsilon$-tolerance, for some user defined sufficiently small $\varepsilon$, and consider the weights to be equal if they are within $\varepsilon$ of each other.

## 9. Semantic Treewidth

A graphical model $\mathcal{M}$ represents a universal function $F = \otimes f_i$. The function $F$ may be represented by different graphical models. Given a particular pseudo tree $\mathcal{T}$, that captures some of the structural information of $F$, we are interested in all the graphical models that accept $\mathcal{T}$ as a pseudo tree, namely their primal graphs only contain edges that are backarcs in $\mathcal{T}$. Since the size of the AOMDD for $F$ based on $\mathcal{T}$ is bounded in the worst case by the induced width of the graphical model along $\mathcal{T}$, we define the *semantic treewidth* to be:

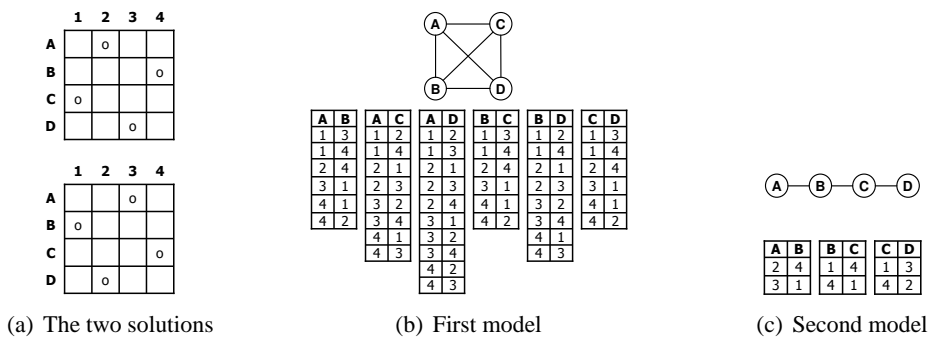

(a) The two solutions        (b) First model        (c) Second model

Figure 23: The 4-queen problem

**DEFINITION 27 (semantic treewidth)** *The semantic treewidth of a graphical model $\mathcal{M}$ relative to a pseudo tree $\mathcal{T}$ denoted by $sw_{\mathcal{T}}(\mathcal{M})$, is the smallest treewidth taken over all models $\mathcal{R}$ that are equivalent to $\mathcal{M}$, and accept the pseudo tree $\mathcal{T}$. Formally, it is defined by $sw_{\mathcal{T}}(\mathcal{M}) = min_{\mathcal{R},u(\mathcal{R})=u(\mathcal{M})} w_{\mathcal{T}}(\mathcal{R})$, where $u(\mathcal{M})$ is the universal function of $\mathcal{M}$, and $w_{\mathcal{T}}(\mathcal{R})$ is the induced width of $\mathcal{R}$ along $\mathcal{T}$. The semantic treewidth of a graphical model, $\mathcal{M}$, is the minimal semantic treewidth over all the pseudo trees that can express its universal function.*

Computing the semantic treewidth can be shown to be NP-hard.[3]

**THEOREM 8** *Computing the semantic treewidth of a graphical model $\mathcal{M}$ is NP-hard.*

Theorem 8 shows that computing the semantic treewidth is hard, and it is likely that the actual complexity is even higher. However, the semantic treewidth can explain why sometimes the minimal AND/OR graph or OBDD are much smaller than the exponential in treewidth or pathwidth upper bounds. In many cases, there could be a huge disparity between the treewidth of $\mathcal{M}$ and the semantic treewidth along $\mathcal{T}$.

**Example 14** *Figure 23(a) shows the two solutions of the 4-queen problem. The problem is expressed by a complete graph of treewidth 3, given in Figure 23(b). Figure 23(c) shows an equivalent problem (i.e., that has the same set of solutions), which has treewidth 1. The semantic treewidth of the 4-queen problem is 1.*

Based on the fact that an AOMDD is a canonical representation of the universal function of a graphical model, we can conclude that the size of the AOMDD is bounded exponentially by the semantic treewidth along the pseudo tree, rather than the treewidth of the given graphical model representation.

**Proposition 7** *The size of the AOMDD of a graphical model $\mathcal{M}$ is bounded by $O(n\ k^{sw_{\mathcal{T}}(\mathcal{M})})$, where $n$ is the number of variables, $k$ is the maximum domain size and $sw_{\mathcal{T}}(\mathcal{M})$ is the semantic treewidth of $\mathcal{M}$ along the pseudo tree $\mathcal{T}$.*

---

3. We thank David Eppstein for the proof.

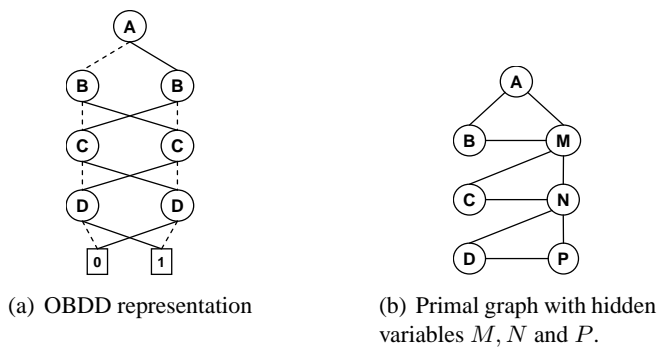

(a) OBDD representation

(b) Primal graph with hidden variables $M, N$ and $P$.

Figure 24: The parity function

**Example 15** *Consider a constraint network on $n$ variables such that every two variables are constrained by equality ($X = Y$). One graph representation is a complete graph, another is a chain and another is a tree. If the problem is specified as a complete graph, and if we use a linear order, the OBDD will have a linear size because there exists a representation having a pathwidth of 1 (rather than $n$).*

While the semantic treewidth can yield a much better upper bound on the AOMDD, it can also be a very bad bound. It is well known that the parity function on $n$ variables has a very compact, chain-like OBDD representation. Yet, the only constraint network representation of a parity function is the function itself (namely a complete graph on all the variables), whose treewidth and semantic treewidth is its number of variables, $n$. The OBDD representation of the parity function suggests that the addition of hidden variables can simplify its presentation. We show an example in Figure 24. On the left side, in Figure 24(a) we have the OBDD representation of the parity function for four binary variables. A graphical model would represent this function by a complete graph on the four variables. However, we could add the extra variables $M, N$ and $P$ in Figure 24(b), sometimes called "hidden" variables, that can help decompose the model. In this case $M$ can form a constraint together with $A$ and $B$ such that $M$ represents the parity of $A$ and $B$, namely $M = 1$ if $A \oplus B = 1$, where $\oplus$ is the parity (XOR) operator. Similarly, $N$ would capture the parity of $M$ and $C$, and $P$ would capture the parity of $N$ and $D$, and would also give the parity of the initial four variables. The two structures are surprisingly similar. It would be interesting to study further the connection between hidden variables and compact AOBDDs, but we leave this for future work.

## 10. Experimental Evaluation

Our experimental evaluation is in preliminary stages, but the results we have are already encouraging. We ran the search-based compile algorithm, by recording the trace of the AND/OR search, and then reducing the resulting AND/OR graph bottom up. In these results we only applied the reduction by isomorphism and still kept the redundant meta-nodes. We implemented our algorithms in C++ and ran all experiments on a 2.2GHz Intel Core 2 Duo with 2GB of RAM, running Windows.

## 10.1 Benchmarks

We tested the performance of the search-based compilation algorithm on random Bayesian networks, instances from the Bayesian Network Repository and a subset of networks from the UAI'06 Inference Evaluation Dataset.

**Random Bayesian Networks**  The random Bayesian networks were generated using parameters $(n, k, c, p)$, where $n$ is the number of variables, $k$ is the domain size, $c$ is the number of conditional probability tables (CPTs) and $p$ is the number of parents in each CPT. The structure of the network was created by randomly picking $c$ variables out of $n$ and, for each, randomly picking $p$ parents from their preceding variables, relative to some ordering. The remaining $n - c$ variables are called *root* nodes. The entries of each probability table were generated randomly using a uniform distribution, and the table was then normalized. It is also possible to control the amount of determinism in the network by forcing a percentage $det$ of the CPTs to have only 0 and 1 entries.

**Bayesian Network Repository**  The Bayesian Network Repository[4] contains a collection of belief networks extracted from various real-life domains which are often used for benchmarking probabilistic inference algorithms.

**UAI'06 Inference Evaluation Dataset**  The UAI 2006 Inference Evaluation Dataset[5] contains a collection of random as well as real-world belief networks that were used during the first UAI 2006 Inference Evaluation contest. For our purpose we selected a subset of networks which were derived from the ISCAS'89 digital circuits benchmark.[6] ISCAS'89 circuits are a common benchmark used in formal verification and diagnosis. Each of these circuits was converted into a Bayesian network by removing flip-flops and buffers in a standard way, creating a deterministic conditional probability table for each gate, and putting uniform distributions on the input signals.

## 10.2 Algorithms

We consider two search-based compilation algorithms, denoted by AOMDD-BCP and AOMDD-SAT, respectively, that reduce the context minimal AND/OR graph explored via isomorphism, while exploiting the determinism (if any) present in the network. The approach we take for handling the determinism is based on unit resolution over a CNF encoding (i.e., propositional clauses) of the zero probability tuples of the CPTs. The idea of using unit resolution during search for Bayesian networks was first explored by Allen and Darwiche (2003). AOMDD-BCP is conservative and applies only unit resolution at each node in the search graph, whereas AOMDD-SAT is more aggressive and detects inconsistency by running a full SAT solver. We used the zChaff SAT solver (Moskewicz, Madigan, Zhao, Zhang, & Malik, 2001) for both unit resolution as well as full satisfiability. For comparison, we also ran an OR version of AOMDD-BCP, called MDD-BCP.

For reference we also report results obtained with the ACE[7] compiler. ACE compiles a Bayesian network into an Arithmetic Circuit (AC) and then uses the AC to answer multiple queries with respect to the network. An arithmetic circuit is a representation that is equivalent to AND/OR graphs (Mateescu & Dechter, 2007). Each time ACE compiler is invoked, it uses one of two algorithms as the basis for compilation. First, if an elimination order can be generated for the network having

---

4. http://www.cs.huji.ac.il/compbio/Repository/

5. http://ssli.ee.washington.edu/bilmes/uai06InferenceEvaluation

6. Available at: http://www.fm.vslib.cz/kes/asic/iscas/

7. Available at: http://reasoning.cs.ucla.edu/ace

| Network | (w*, h) | (n, k) | ACE | | MDD w/ BCP | | | AOMDD w/ BCP | | | AOMDD w/ SAT | | |
|---|---|---|---|---|---|---|---|---|---|---|---|---|---|
| | | | #nodes | time | #meta | #cm(OR) | time | #meta | #cm(OR) | time | #meta | #cm(OR) | time |
| Bayesian Network Repository | | | | | | | | | | | | | |
| alarm | (4, 13) | (37, 4) | 1,511 | 0.01 | 208,837 | 682,195 | 73.35 | **320** | 459 | 0.05 | **320** | 459 | 0.22 |
| cpcs54 | (14, 23) | (54, 2) | 196,933 | 0.06 | - | - | - | 65,158 | 66,405 | 6.97 | 65,158 | 66,405 | 6.97 |
| cpcs179 | (8, 14) | (179, 4) | 67,919 | 0.05 | - | - | - | 9,990 | 32,185 | 46.56 | 9,990 | 32,185 | 46.56 |
| cpcs360b | (20, 27) | (360, 2) | **5,258,826** | 1.72 | - | - | - | - | - | - | - | - | - |
| diabetes | (4, 77) | (413, 21) | **7,615,989** | 1.81 | - | - | - | - | - | - | - | - | - |
| hailfinder | (4, 16) | (56, 11) | 8,815 | 0.01 | - | - | - | 2,068 | 2,202 | 0.34 | **1,893** | 2,202 | 1.48 |
| mildew | (4, 13) | (35, 100) | 823,913 | 0.39 | - | - | - | 73,666 | 110,284 | 1367.81 | **62,903** | 65,599 | 3776.82 |
| mm | (20, 57) | (1220, 2) | 47,171 | 1.49 | - | - | - | 38,414 | 58,144 | 4.54 | **30,274** | 52,523 | 99.55 |
| munin2 | (9, 32) | (1003, 21) | 2,128,147 | 1.91 | - | - | - | - | - | - | - | - | - |
| munin3 | (9, 32) | (1041, 21) | 1,226,635 | 1.27 | - | - | - | - | - | - | - | - | - |
| munin4 | (9, 32) | (1044, 21) | 2,423,009 | 4.44 | - | - | - | - | - | - | - | - | - |
| pathfinder | (6, 11) | (109, 63) | 18,250 | 0.05 | 610,854 | 1,303,682 | 352.18 | 6,984 | 16,267 | 30.71 | **2,265** | 15,963 | 50.36 |
| pigs | (11, 26) | (441, 3) | 636,684 | 0.19 | - | - | - | 261,920 | 294,101 | 174.29 | **198,284** | 294,101 | 1277.72 |
| water | (10, 15) | (32, 4) | 59,642 | 0.52 | 707,283 | 1,138,096 | 95.14 | 18,744 | 20,926 | 2.02 | **18,503** | 19,225 | 7.45 |
| UAI'06 Evaluation Dataset | | | | | | | | | | | | | |
| BN_42 | (21, 62) | (851, 2) | **4,860** | 1.35 | - | - | - | 107,025 | 341,428 | 53.50 | 42,445 | 43,280 | 57.36 |
| BN_43 | (26, 65) | (851, 2) | **10,373** | 1.62 | - | - | - | 1,343,923 | 1,679,013 | 1807.63 | 313,388 | 314,669 | 434.38 |
| BN_44 | (25, 56) | (851, 2) | **4,235** | 1.31 | - | - | - | 155,588 | 187,589 | 20.90 | 47,222 | 48,540 | 66.09 |
| BN_45 | (22, 54) | (851, 2) | **12,319** | 1.50 | - | - | - | 390,795 | 487,593 | 68.81 | 126,182 | 126,929 | 177.50 |
| BN_46 | (20, 46) | (851, 2) | **5,912** | 2.90 | 1,125,658 | 1,228,332 | 94.93 | 16,711 | 17,532 | 1.31 | 7,337 | 7,513 | 5.54 |
| BN_47 | (39, 57) | (632, 2) | 1,448 | 1.17 | 42,419 | 47,128 | 2.87 | 1,873 | 2,663 | 0.24 | **1,303** | 2,614 | 2.36 |
| BN_49 | (40, 60) | (632, 2) | 1,408 | 1.16 | 18,344 | 19,251 | 1.32 | 1,205 | 1,539 | 0.19 | **952** | 1,515 | 1.34 |
| BN_51 | (41, 68) | (632, 2) | **1,467** | 1.15 | 63,851 | 68,005 | 4.22 | 4,442 | 5,267 | 0.50 | 3,653 | 5,195 | 4.58 |
| BN_53 | (47, 87) | (532, 2) | **1,357** | 0.91 | 14,210 | 19,162 | 1.49 | 4,819 | 9,561 | 0.74 | 1,365 | 1,719 | 1.36 |
| BN_55 | (49, 92) | (532, 2) | 1,288 | 0.93 | 5,168 | 6,088 | 0.57 | 1,972 | 2,816 | 0.26 | **790** | 904 | 0.75 |
| BN_57 | (49, 85) | (532, 2) | 1,276 | 0.90 | 48,436 | 51,611 | 3.52 | 4,036 | 5,089 | 0.37 | **962** | 1,277 | 1.01 |
| BN_59 | (52, 87) | (511, 2) | **1,749** | 0.93 | 332,030 | 353,720 | 25.61 | 22,963 | 29,146 | 2.14 | 10,655 | 18,752 | 14.17 |
| BN_61 | (41, 64) | (638, 2) | 1,411 | 1.10 | 20,459 | 20,806 | 1.45 | 1,244 | 1,589 | 0.17 | **1,016** | 1,528 | 1.37 |
| BN_63 | (53, 95) | (511, 2) | **1,324** | 0.90 | 11,461 | 17,087 | 1.28 | 7,182 | 14,048 | 1.07 | 1,419 | 2,177 | 1.69 |
| BN_65 | (56, 86) | (411, 2) | **1,184** | 0.75 | - | - | - | 20,764 | 23,102 | 1.52 | 12,569 | 19,778 | 12.90 |
| BN_67 | (54, 88) | (411, 2) | 1,031 | 0.74 | - | - | - | 179,067 | 511,031 | 154.91 | **716** | 1,169 | 0.78 |
| Positive Random Bayesian Networks (n=75, k=2, p=2, c=65) | | | | | | | | | | | | | |
| r75-1 | (12, 22) | (75, 2) | 67,737 | 0.31 | - | - | - | **21,619** | 21,619 | 2.59 | **21,619** | 21,619 | 2.59 |
| r75-2 | (12, 23) | (75, 2) | 46,703 | 0.29 | - | - | - | **18,083** | 18,083 | 1.88 | **18,083** | 18,083 | 1.88 |
| r75-3 | (11, 26) | (75, 2) | 53,245 | 0.30 | - | - | - | **18,419** | 18,419 | 1.86 | **18,419** | 18,419 | 1.86 |
| r75-4 | (11, 19) | (75, 2) | 28,507 | 0.29 | - | - | - | **8,363** | 8,363 | 1.16 | **8,363** | 8,363 | 1.16 |
| r75-5 | (13, 24) | (75, 2) | 149,707 | 0.36 | - | - | - | **42,459** | 42,459 | 4.61 | **42,459** | 42,459 | 4.61 |
| r75-6 | (14, 24) | (75, 2) | 132,107 | 1.19 | - | - | - | **62,621** | 62,621 | 6.95 | **62,621** | 62,621 | 6.95 |
| r75-7 | (12, 24) | (75, 2) | 89,913 | 0.36 | - | - | - | **21,583** | 21,583 | 2.42 | **21,583** | 21,583 | 2.42 |
| r75-8 | (14, 24) | (75, 2) | 86,183 | 0.36 | - | - | - | **49,001** | 49,001 | 6.23 | **49,001** | 49,001 | 6.23 |
| r75-9 | (11, 19) | (75, 2) | 29,025 | 0.30 | - | - | - | **7,681** | 7,681 | 0.81 | **7,681** | 7,681 | 0.81 |
| r75-10 | (10, 24) | (75, 2) | 20,291 | 0.28 | - | - | - | **5,905** | 5,905 | 0.63 | **5,905** | 5,905 | 0.63 |
| Deterministic Random Bayesian Networks (n=100, k=2, p=2, c=90) and det = 25% of the CPTs containing only 0 and 1 entries | | | | | | | | | | | | | |
| r100d25-1 | (13, 31) | (100, 2) | 68,398 | 0.38 | - | - | - | **34,035** | 34,075 | 2.94 | **34,035** | 34,075 | 12.77 |
| r100d25-2 | (16, 28) | (100, 2) | 150,134 | 0.46 | - | - | - | **70,241** | 70,931 | 7.72 | **70,241** | 70,931 | 27.17 |
| r100d25-3 | (16, 29) | (100, 2) | 705,200 | 0.96 | - | - | - | **134,079** | 135,203 | 13.80 | **134,079** | 135,203 | 50.51 |
| r100d25-4 | (16, 31) | (100, 2) | 161,902 | 0.54 | - | - | - | **79,366** | 79,488 | 7.26 | **79,366** | 79,488 | 28.06 |
| r100d25-5 | (16, 29) | (100, 2) | 185,348 | 0.53 | - | - | - | **140,627** | 140,636 | 14.57 | **140,627** | 140,636 | 49.42 |
| r100d25-6 | (18, 28) | (100, 2) | **148,835** | 0.66 | - | - | - | 204,232 | 210,066 | 17.56 | 197,134 | 210,066 | 92.24 |
| r100d25-7 | (16, 29) | (100, 2) | 264,629 | 0.60 | - | - | - | 134,344 | 135,008 | 14.26 | **133,850** | 135,008 | 55.60 |
| r100d25-8 | (17, 27) | (100, 2) | 65,186 | 0.46 | - | - | - | **36,857** | 36,887 | 2.95 | **36,857** | 36,887 | 11.97 |
| r100d25-9 | (14, 27) | (100, 2) | 140,014 | 0.40 | - | - | - | 58,421 | 59,791 | 6.88 | **58,172** | 59,791 | 23.21 |
| r100d25-10 | (16, 27) | (100, 2) | 173,808 | 0.58 | - | - | - | **69,110** | 69,136 | 7.50 | **69,110** | 69,136 | 26.50 |

Table 1: Results for experiments with 50 Bayesian networks from 3 problem classes; $w^*$ = treewidth, $h$ = depth of pseudo tree, $n$ = number of variables, $k$ = domain size, *time* given in seconds; bold types highlight the best results across rows.

sufficiently small induced width, then tabular variable elimination will be used as the basis. This algorithm is similar to the one discussed by Chavira and Darwiche (2007), but uses tables to represent factors rather than ADDs. If the induced width is large, then logical model counting will be used as the basis. Tabular variable elimination is typically efficient when width is small but cannot handle networks when the width is larger. Logical model counting, on the other hand, incurs more overhead than tabular variable elimination, but can handle many networks having larger treewidth. Both tabular variable elimination and logical model counting produce ACs that exploit local structure, leading to efficient online inference. When logical model counting is invoked, it proceeds by encoding the Bayesian network into a CNF (Chavira & Darwiche, 2005; Chavira, Darwiche, & Jaeger, 2006), simplifying the CNF, compiling the CNF into a d-DNNF, and then extracting the AC from the compiled d-DNNF. A dtree over the CNF clauses drives the compilation step.

In all our experiments we report the compilation time in seconds ($time$), the number of OR nodes in the context minimal graph explored ($\#cm$), the number of meta-nodes of the resulting AOMDD ($\#meta$), as well as the size of the AC compiled by ACE ($\#nodes$). For each network we specify the number of variables ($n$), domain size ($k$), induced width ($w^*$) and pseudo tree depth ($h$). A '-' stands for exceeding the 2GB memory limit by the respective algorithm. The best performance points are highlighted.

## 10.3 Evaluation on Bayesian Networks

Table 1 reports the results obtained for experiments with 50 Bayesian networks. The AOMDD compilers as well as ACE used the min-fill heuristic (Kjæaerulff, 1990) to construct the guiding pseudo tree and dtree, respectively.

### 10.3.1 BAYESIAN NETWORKS REPOSITORY

We see that ACE is overall the fastest compiler on this domain, outperforming both AOMDD-BCP and AOMDD-SAT with up to several orders of magnitude (e.g., `mildew`, `pigs`). However, the diagrams compiled by ACE and AOMDD-BCP (resp. AOMDD-SAT) are comparable in size. In some cases, AOMDD-BCP and AOMDD-SAT were able to compile much smaller diagrams than ACE. For example, the diagram produced by AOMDD-BCP for the `mildew` network is 13 times smaller than the one compiled by ACE. In principle the output produced by ACE and AOMDD should be similar if both are guided by the same pseudo tree/dtree. Our scheme should be viewed as a compilation alternative which (1) extends decision diagrams and (2) mimics traces of search properties that may make this representation accessible. The OR compiler MDD-BCP was able to compile only 3 out of the 14 test instances, but their sizes were far larger than those produced by AOMDD-BCP. For instance, on the `pathfinder` network, AOMDD-BCP outputs a decision diagram almost 2 orders of magnitude smaller than MDD-BCP.

### 10.3.2 UAI'06 DATASET

For each of the UAI'06 Dataset instances we picked randomly 30 variables and instantiated as evidence. We see that ACE is the best performing compiler on this dataset. AOMDD-BCP is competitive with ACE in terms of compile time only on 9 out the 16 test instances. AOMDD-SAT is able to compile the smallest diagrams for 6 networks only (e.g., `BN_47`, `BN_49`, `BN_55`, `BN_57`, `BN_61`, `BN_67`). As before, the difference in size between the compiled data-structures produces by MDD-BCP and AOMDD-BCP is up to 2 orders of magnitude in favor of the latter.

### 10.3.3 Random Networks

The problem instances denoted by *r75-1* through *r75-10* were generated from a class of random belief networks with parameters ($n = 75, k = 2, p = 2, c = 65$). Similarly, the instances denoted by *r100d25-1* through *r100d25-10* belong to a class with parameters ($n = 100, k = 2, p = 2, c = 90$). In the latter case, $det = 25\%$ of the CPTs are deterministic, namely they contain only 0 and 1 probability tuples. These test instances were compiled without any evidence. We see that on this domain AOMDD-BCP/AOMDD-SAT were able to compile the smallest diagrams, which were on average about 2 times smaller than those produced by ACE. However, ACE was again the fastest compiler. Notice that the OR compiler MDD-BCP ran out of memory in all test cases.

### 10.4 The Impact of Variable Ordering

As theory dictates, the AOMDD size is influenced by the quality of the guiding pseudo tree. In addition to the min-fill heuristic we also considered the *hypergraph* heuristic which constructs the pseudo tree by recursively decomposing the dual hypergraph associated with the graphical model. This idea was also explored by Darwiche (2001) for constructing dtrees that guide ACE.

Since both the min-fill and hypergraph partitioning heuristics are randomized (namely ties are broken randomly), the size of the AOMDD guided by the resulting pseudo tree may vary significantly from one run to the next. Figure 25 displays the AOMDD size using hypergraph and min-fill based pseudo trees for 6 networks selected from Table 1, over 20 independent runs. We also record the average induced width and depth obtained for the pseudo trees (see the header of each plot in Figure 25). We see that the two heuristics do not dominate each other, namely the variance in output size is quite significant in both cases.

### 10.5 Memory Usage

Table 2 shows the memory usage (in MBytes) of ACE, AOMDD-BCP and AOMDD-SAT, respectively, on the Bayesian networks from Table 1. We see that in some cases the AOMDD based compilers require far less memory than ACE. For example, on the "mildew" network, both AOMDD-BCP and AOMDD-SAT use about 22 MB of memory to compile the AND/OR decision diagram, while ACE requires as much as 218 MB of memory. Moreover, the compiled AOMDD has in this case about one order of magnitude fewer nodes than that constructed by ACE. When comparing the two AND/OR search-based compilers, we observe that on networks with a significant amount of determinism, such as those from the UAI'06 Evaluation dataset, AOMDD-SAT uses on average two times less memory than AOMDD-BCP. The most dramatic savings in memory usage due to the aggressive constraint propagation employed by AOMDD-SAT compared with AOMDD-BCP can be seen on the "BN_67" network. In this case, the difference in memory usage between AOMDD-SAT and AOMDD-BCP is about 2 orders of magnitude in favor of the former.

## 11. Related Work

The related work can be viewed along two directions: (1) the work related to the AND/OR search idea for graphical models and (2) the work related to compilation for graphical models that exploits problem structure.

An extensive discussion for (1) was provided in the previous work of Dechter and Mateescu (2007). Since this is not the focus of the paper, we just mention that the AND/OR idea was origi-nally developed for heuristic search (Nilsson, 1980). As mentioned in the introduction, the AND/OR search for graphical models is based on a pseudo tree that spans the graph of the model, similar to the tree rearrangement of Freuder and Quinn (1985, 1987). The idea was adapted for distributed constraint satisfaction by Collin et al. (1991, 1999) and more recently by Modi et al. (2005), and was also shown to be related to graph-based backjumping (Dechter, 1992). This work was extended by Bayardo and Miranker (1996), Bayardo and Schrag (1997) and more recently applied to optimization tasks by Larrosa et al. (2002). Another version that can be viewed as exploring the AND/OR graphs was presented recently for constraint satisfaction (Terrioux & Jégou, 2003b) and for optimization (Terrioux & Jégou, 2003a). Similar principles were introduced recently for probabilistic inference, in algorithm Recursive Conditioning (Darwiche, 2001) as well as in Value Elimination (Bacchus et al., 2003b, 2003a), and are currently at the core of the most advanced SAT solvers (Sang et al., 2004).

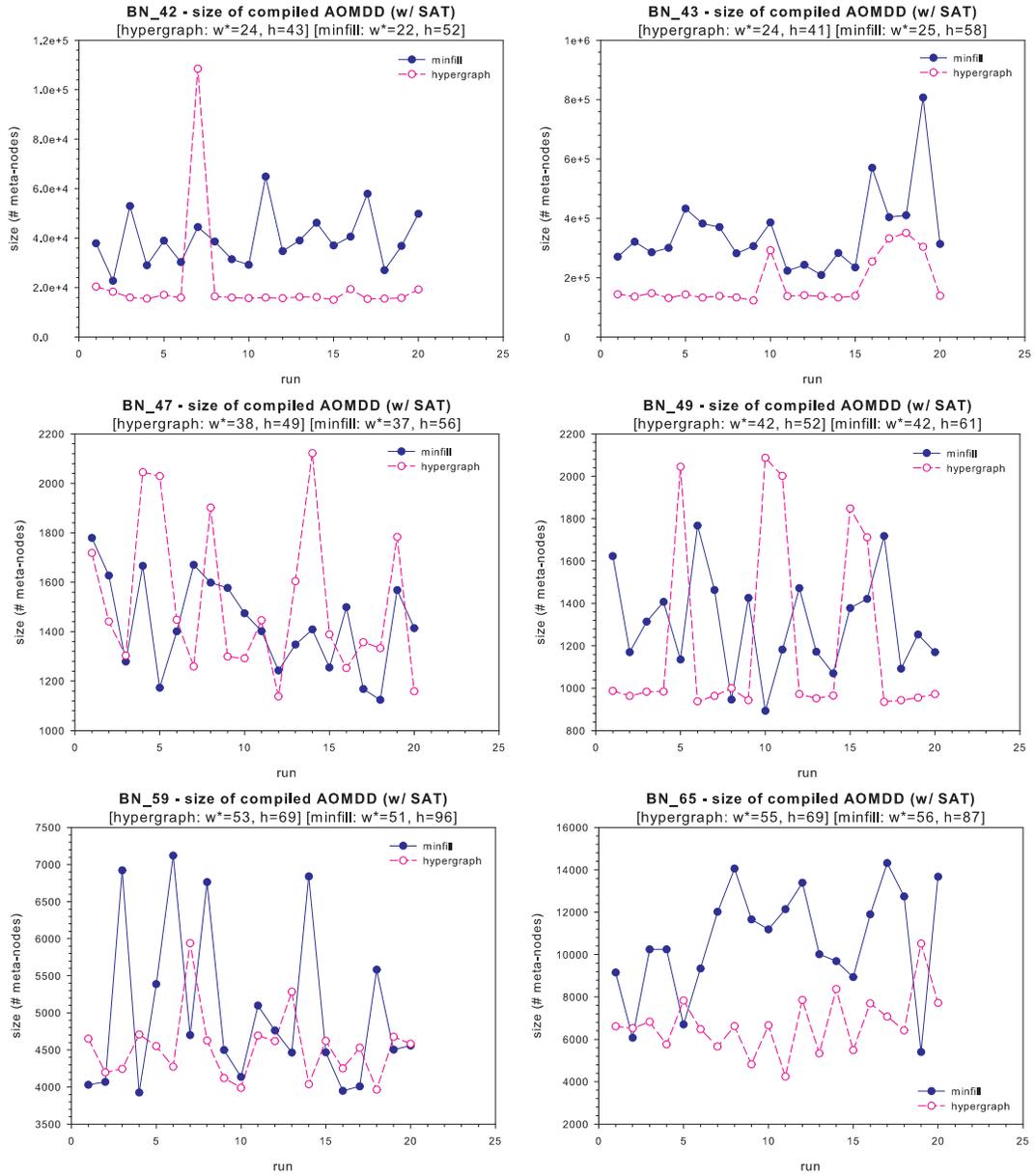

Figure 25: Effect of variable ordering.

| Network | ACE | | AOMDD w/ BCP | | AOMDD w/ SAT | |
|---|---|---|---|---|---|---|
| | #nodes | memory (MB) | #nodes | memory (MB) | #nodes | memory (MB) |
| Bayesian Network Repository | | | | | | |
| alarm | 1,511 | **0.00** | 320 | 0.0206 | 320 | 0.0206 |
| cpcs54 | 196,933 | 4.00 | 65,158 | **3.4415** | 65,158 | **3.4415** |
| cpcs179 | 67,919 | 5.00 | 9,990 | **1.9263** | 9,990 | **1.9263** |
| cpcs360b | 5,258,826 | **204.00** | - | - | - | - |
| diabetes | 7,615,989 | **449.00** | - | - | - | - |
| hailfinder | 8,815 | **0.00** | 2,068 | 0.1576 | 1,893 | 0.1740 |
| mildew | 823,913 | 218.00 | 73,666 | **22.5781** | 62,903 | **22.1467** |
| mm | 47,171 | 369.00 | 38,414 | 1.5719 | 30,274 | **1.3711** |
| munin2 | 2,128,147 | **202.00** | - | - | - | - |
| munin3 | 1,226,635 | **150.00** | - | - | - | - |
| munin4 | 2,423,009 | n/a | - | - | - | - |
| pathfinder | 18,250 | 10.00 | 6,984 | 0.6009 | 2,265 | **0.3515** |
| pigs | 636,684 | 31.00 | 261,920 | 23.3761 | 198,284 | **17.7096** |
| water | 59,642 | 161.00 | 18,744 | **1.09578** | 18,503 | 1.3258 |
| UAI'06 Evaluation Dataset | | | | | | |
| BN_42 | 4,860 | n/a | 107,025 | 4.5622 | 42,445 | **1.9323** |
| BN_43 | 10,373 | n/a | 1,343,923 | 57.8422 | 313,388 | **14.2828** |
| BN_44 | 4,235 | n/a | 155,588 | 6.5613 | 47,222 | **2.1628** |
| BN_45 | 12,319 | n/a | 390,795 | 17.9325 | 126,182 | **5.7958** |
| BN_46 | 5,912 | n/a | 16,711 | 0.6929 | 7,337 | **0.3401** |
| BN_47 | 1,448 | n/a | 1,873 | 0.0720 | 1,303 | **0.0583** |
| BN_49 | 1,408 | n/a | 1,205 | 0.0449 | 952 | **0.0409** |
| BN_51 | 1,467 | n/a | 4,442 | 0.1689 | 3,653 | **0.1633** |
| BN_53 | 1,357 | n/a | 4,819 | 0.1814 | 1,365 | **0.0587** |
| BN_55 | 1,288 | n/a | 1,972 | 0.0723 | 790 | **0.0336** |
| BN_57 | 1,276 | n/a | 4,036 | 0.1495 | 962 | **0.0411** |
| BN_59 | 1,749 | n/a | 22,963 | 0.8501 | 10,655 | **0.4587** |
| BN_61 | 1,411 | n/a | 1,244 | 0.0463 | 1,016 | **0.0445** |
| BN_63 | 1,324 | n/a | 7,182 | 0.2728 | 1,419 | **0.0607** |
| BN_65 | 1,184 | n/a | 20,764 | 0.7539 | 12,569 | **0.5384** |
| BN_67 | 1,031 | n/a | 179,067 | 6.9603 | 716 | **0.0304** |
| Positive Random Bayesian Networks with parameters (n=75, k=2, p=2, c=65) | | | | | | |
| r75-1 | 67,737 | **1.00** | 21,619 | 1.2503 | 21,619 | 1.2503 |
| r75-2 | 46,703 | 1.00 | 18,083 | **0.9957** | 18,083 | **0.9957** |
| r75-3 | 53,245 | 1.00 | 18,419 | **0.9955** | 18,419 | **0.9955** |
| r75-4 | 28,507 | 1.00 | 8,363 | **0.5171** | 8,363 | **0.5171** |
| r75-5 | 149,707 | 3.00 | 42,459 | **2.3299** | 42,459 | **2.3299** |
| r75-6 | 132,107 | **3.00** | 62,621 | 3.4330 | 62,621 | 3.4330 |
| r75-7 | 89,913 | 2.00 | 21,583 | **1.1942** | 21,583 | **1.1942** |
| r75-8 | 86,183 | **2.00** | 49,001 | 2.8130 | 49,001 | 2.8130 |
| r75-9 | 29,025 | 1.00 | 7,681 | **0.4124** | 7,681 | **0.4124** |
| r75-10 | 20,291 | 1.00 | 5,905 | **0.3261** | 5,905 | **0.3261** |
| Deterministic Random Bayesian Networks with parameters (n=100, k=2, p=2, c=90) | | | | | | |
| r100d25-1 | 68,398 | 5.00 | 34,035 | **1.6290** | 34,035 | 1.7149 |
| r100d25-2 | 150,134 | 10.00 | 70,241 | **3.6129** | 70,241 | 3.7810 |
| r100d25-3 | 705,200 | 40.00 | 134,079 | **6.6372** | 134,079 | 6.9873 |
| r100d25-4 | 161,902 | 22.00 | 79,366 | **3.8113** | 79,366 | 4.0079 |
| r100d25-5 | 185,348 | 15.00 | 140,627 | **7.0839** | 140,627 | 7.4660 |
| r100d25-6 | 148,835 | 37.00 | 204,232 | **9.1757** | 197,134 | 9.6542 |
| r100d25-7 | 264,629 | 19.00 | 134,344 | **6.9619** | 133,850 | 6.9961 |
| r100d25-8 | 65,186 | 21.00 | 36,857 | **1.6872** | 36,857 | 1.8278 |
| r100d25-9 | 140,014 | 6.00 | 58,421 | **3.1058** | 58,172 | 3.2055 |
| r100d25-10 | 173,808 | 27.00 | 69,110 | **3.5578** | 69,110 | 3.6636 |

Table 2: Memory usage in MBytes of ACE, AOMDD-BCP and AOMDD-SAT on the 50 Bayesian networks from Table 1. Bold types highlight the best performance across rows. The "n/a" indicates that the respective memory usage statistic was not available from ACE's output.

For direction (2), there are various lines of related research. The formal verification literature, beginning with the work of Bryant (1986) contains a very large number of papers dedicated to the study of BDDs. However, BDDs are in fact OR structures (the underlying pseudo tree is a chain) and do not take advantage of the problem decomposition in an explicit way. The complexity bounds for OBDDs are based on *pathwidth* rather than *treewidth*.

As noted earlier, the work of Bertacco and Damiani (1997) on Disjoint Support Decomposition (DSD) is related to AND/OR BDDs in various ways. The main common aspect is that both approaches show how structure decomposition can be exploited in a BDD-like representation. DSD is focused on Boolean functions and can exploit more refined structural information that is inherent to Boolean functions. In contrast, AND/OR BDDs assume only the structure conveyed in the constraint graph, and are therefore more broadly applicable to any constraint expression and also to graphical models in general. They allow a simpler and higher level exposition that yields graph-based bounds on the overall size of the generated AOMDD. The full relationship between these two formalisms should be studied further.

McMillan (1994) introduced the BDD trees, along with the operations for combining them. For circuits of bounded tree width, BDD trees have a linear space upper bound of $O(|g|2^{w2^{2w}})$, where $|g|$ is the size of the circuit $g$ (typically linear in the number of variables) and $w$ is the treewidth. This bound hides some very large constants to claim the linear dependence on $|g|$ when $w$ is bounded. However, McMillan maintains that when the input function is a CNF expression BDD-trees have the same bounds as AND/OR BDDs, namely they are exponential in the treewidth only.

To sketch just a short comparison between McMillan's BDD trees and AOMMDs, consider an example where we have a simple pseudo tree with root $\alpha$, left child $\beta$ and right child $\gamma$. Each of these nodes may stand for a set of variables. In BDD trees, the assignments to $\beta$ are grouped into equivalence classes according to the cofactors generated by them on the remaining $\alpha$ and $\gamma$. For example assignments $\beta_1$ and $\beta_2$ are equivalent if they generate the same function on $\alpha$ and $\gamma$. The node $\beta$ can be represented by a BDD whose leaves are the cofactors. The same is done for $\gamma$. The node $\alpha$ is then represented by a matrix of BDDs, where each column corresponds to a cofactor of $\beta$ and each line to a cofactor of $\gamma$. By contrast, an AOMDD represents the node $\alpha$ as a BDD whose leaves are the cofactors (the number of distinct functions on $\beta$ and $\gamma$) and then each cofactor is the root of a decomposition (an AND node) between $\beta$ and $\gamma$. Moreover, the representations of $\beta$ (as descendants of different cofactor of $\alpha$) are shared as much as possible and the same goes for $\gamma$. This is only a high level description, that becomes slightly more complicated when redundant nodes are eliminated, but the idea remains the same.

The AND/OR structure restricted to propositional theories is very similar to deterministic decomposable negation normal form (d-DNNF) (Darwiche & Marquis, 2002; Darwiche, 2002). More recently, Huang and Darwiche (2005b) used the trace of the DPLL algorithm to generate an OBDD, and compared with the typical formal verification approach of combining the OBDDs of the input function according to some schedule. The structures that were investigated in that case are still OR. This idea is extended in our present work by the AND/OR search compilation algorithm.

McAllester, Collins, and Pereira (2004) introduced the case factor diagrams (CFD), which subsume Markov random fields of bounded tree width and probabilistic context free grammars (PCFG). CFDs are very much related to the AND/OR graphs. The CFDs target the minimal representation, by exploiting decomposition (similar to AND nodes) but also by exploiting context sensitive information and allowing dynamic ordering of variables based on context. CFDs do not eliminate the redundant nodes, and part of the cause is that they use zero suppression. There is no claim about CFDs being canonical forms, and also no description of how to combine two CFDs.

There are numerous variants of decision diagrams that are designed to represent integer-valued or real-valued functions. For a comprehensive view we refer the reader to the survey of Drechsler and Sieling (2001). Algebraic decision diagrams (ADDs) (Bahar et al., 1993) provide a compilation for general real-valued rather than Boolean functions. Their main drawback is that their size increases very fast if the number of terminals becomes large. There are several approaches that try to alleviate this problem. However the structure that they capture is still OR, and they do not exploit decomposition. Some alternatives introduce edge values (or weights) that enable more subgraph sharing. Edge-valued binary decision diagrams (EVBDDs) (Lai & Sastry, 1992) use additive weights, and when multiplicative weights are also allowed they are called factored EVBDDs (FEVBDDs) (Tafertshofer & Pedram, 1997). Another type of BDDs called K*BMDs (Drechsler, Becker, & Ruppertz, 1996) also use integer weights, both additive and multiplicative in parallel. ADDs have also been extended to affine ADDs (Sanner & McAllester, 2005), through affine transformations that can achieve more compression. The result was shown to be beneficial for probabilistic inference algorithms, such as tree clustering, but they still do not exploit the AND structure.

More recently, independently and in parallel to our work on AND/OR graphs (Dechter & Mateescu, 2004a, 2004b), Fargier and Vilarem (2004) and Fargier and Marquis (2006, 2007) proposed the compilation of CSPs into tree-driven automata, which have many similarities to our work. Their main focus is the transition from linear automata to tree automata (similar to that from OR to AND/OR), and the possible savings for tree-structured networks and hyper-trees of constraints due to decomposition. Their compilation approach is guided by a tree-decomposition while ours is guided by a variable-elimination based algorithms. And it is well known that Bucket Elimination and cluster-tree decomposition are in principle the same (Dechter & Pearl, 1989).

Wilson (2005) extended OBDDs to semi-ring BDDs. The semi-ring treatment is restricted to the OR search spaces, but allows dynamic variable ordering. It is otherwise very similar in aim and scope to our AOMDD. When restricting the AOMDD to OR graphs only, the two are closely related, except that we express BDDs using the Shenoy-Shafer axiomatization that is centered on the two operation of combination and marginalization rather then on the semi-ring formulation. Minimality in the formulation of Wilson (2005) is more general allowing merging nodes having different values and therefore it can capture symmetries (called interchangeability).

Another framework very similar to AOMDDs, that we became aware of only recently, is Probabilistic Decision Graphs (PDG) of Jaeger (2004). This work preceded most of the relevant work we discussed above (Fargier & Vilarem, 2004; Wilson, 2005) and went somewhat unnoticed, perhaps due to notational and cultural differences. It is however similar in motivation, framework and proposed algorithms. We believe our AND/OR framework is more accessible. We define the framework over multi-valued domains, provide greater details in algorithms and complexity analysis, make an explicit connection with search frameworks, fully address the issues of canonicity as well as provide an empirical demonstration. In particular, the claim of canonicity for PDGs is similar to the one we make for AOMDDs of weighted models, in that it is relative to the trees (or forests) that can represent the given probability distribution.

There is another line of research by Drechsler and his group (e.g. Zuzek, Drechsler, & Thornton, 2000), who use AND/OR graphs for Boolean function representation, that may seem similar to our approach. However, the semantics and purpose of their AND/OR graphs are different. They are constructed based on the technique of recursive learning and are used to perform Boolean reasoning, i.e. to explore the logic consequences of a given assumption based on the structure of the circuit, especially to derive sets of implicants. The meaning of AND and OR in their case is related to the meaning of the gates/functions, while in our case the meaning is not related to the semantic of the functions. The AND/OR enumeration tree that results from a circuit according to Zuzek et al. (2000) is not related to the AND/OR decomposition that we discuss.

## 12. Conclusion

We propose the AND/OR multi-valued decision diagram (AOMDD), which emerges from the study of AND/OR search spaces for graphical models (Dechter & Mateescu, 2004a, 2004b; Mateescu & Dechter, 2005; Dechter & Mateescu, 2007) and ordered binary decision diagrams (OBDDs) (Bryant, 1986). This data-structure can be used to compile any graphical model.

Graphical models algorithms that are search-based and compiled data-structures such as BDDs differ primarily by their choices of time vs. memory. When we move from regular OR search space to an AND/OR search space the spectrum of algorithms available is improved for all time vs. memory decisions. We believe that the AND/OR search space clarifies the available choices and helps guide the user into making an informed selection of the algorithm that would fit best the particular query asked, the specific input function and the available computational resources.

The contribution of our work is: (1) We formally describe the AOMDD and prove that it is a canonical representation of a constraint network. (2) We extend the AOMDD to general weighted graphical models. (3) We give a compilation algorithm based on AND/OR search, that saves the trace of a memory intensive search (the context minimal AND/OR graph), and then reduces it in one bottom up pass. (4) We describe the APPLY operator that combines two AOMDDs by an operation and show that its complexity is quadratic in the input, but never worse than exponential in the treewidth. (5) We give a scheduling order for building the AOMDD of a graphical model starting with the AOMDDs of its functions which is based on a Variable Elimination algorithm. This guarantees that the complexity is at most exponential in the *induced width* (treewidth) along the ordering. (6) We show how AOMDDs relate to various earlier and recent compilation frameworks, providing a unifying perspective for all these methods. (7) We introduce the *semantic treewidth*, which helps explain why compiled decision diagrams are often much smaller than the worst case bound. Finally, (8) we provide a preliminary empirical demonstration of the power of the current scheme.

## Acknowledgments

This work was done while Robert Mateescu and Radu Marinescu were at the University of California, Irvine. The authors would like to thank the anonymous reviewers for their constructive suggestions to improve the paper, David Eppstein for a useful discussion of complexity issues, and Lars Otten and Natasha Flerova for comments on the final version of the manuscript. This work was supported by the NSF grants IIS-0412854 and IIS-0713118, and the initial part by the Radcliffe fellowship 2005-2006 (through the partner program), with Harvard undergraduate student John Cobb.

## Appendix

### Proof of Proposition 1
Consider the level of variable $X_i$, and the meta-nodes in the list $L^{X_i}$. After one pass through the meta-nodes in $L^{X_i}$ (the inner for loop), there can be no two meta-nodes at the level of $X_i$ in the AND/OR graph that are isomorphic, because they would have been merged in line 6. Also, during the same pass through the meta-nodes in $L^{X_i}$ all the redundant meta-nodes in $L^{X_i}$ are eliminated in line 8. Processing the meta-nodes in the level of $X_i$ will not create new redundant or isomorphic meta-nodes in the levels that have been processed before. It follows that the resulting AND/OR graph is completely reduced. □

### Proof of Theorem 4
The bound on the size follows directly from Theorem 3. The AOMDD size can only be smaller than the size of the context minimal AND/OR graph, which is bounded by $O(n\ k^{w_T^*(G)})$. To prove the time bound, we have to rely on the use of the hash table, and the assumption that an efficient implementation allows an access time that is constant. The time bound of AND/OR-SEARCH-AOMDD is $O(n\ k^{w_T^*(G)})$, from Theorem 3, because it takes time linear in the output (we assume here that no constraint propagation is performed during search). Procedure BOTTOMUPREDUCTION (procedure 1) takes time linear in the size of the context minimal AND/OR graph. Therefore, the AOMDD can be computed in time $O(n\ k^{w_T^*(G)})$, and the result is the same for the algorithm that performs the reduction during the search. □

### Proof of Proposition 2
The complexity of OBDD (and MDD) *apply* is known to be quadratic in the input. Namely, the number of nodes in the output is at most the product of number of nodes in the input. Therefore, the number of nodes that can appear along one path in the output AOMDD can be at most the product of the number of nodes in each input, along the same path, $|\mathcal{G}_f^i| \cdot |\mathcal{G}_g^i|$. Summing over all the paths in $\mathcal{T}$ gives the result. □

### Proof of Proposition 3
The argument is identical to the case of MDDs. The recursive calls in APPLY lead to combinations of one node from $\mathcal{G}_f^{aomdd}$ and one node from $\mathcal{G}_g^{aomdd}$ (rather than a list of nodes). The number of total possible such combinations is $O(|\mathcal{G}_f^{aomdd}| \cdot |\mathcal{G}_g^{aomdd}|)$. □

### Proof of Proposition 4
The recursive calls of APPLY can generate one meta-node in the output for each combination of nodes from $\mathcal{G}_f^{f \cap g}$ and $\mathcal{G}_g^{f \cap g}$. Let's look at combinations of nodes from $\mathcal{G}_f^{f \cap g}$ and $\mathcal{G}_g^{aomdd} \setminus \mathcal{G}_g^{f \cap g}$. The meta-nodes from $\mathcal{G}_g^{aomdd} \setminus \mathcal{G}_g^{f \cap g}$ that can participate in such combinations (let's call this set $\mathcal{A}$) are only those from levels (of variables) right below $\mathcal{T}_{f \cap g}$. This is because of the mechanics of the recursive calls in APPLY. Whenever a node from $f$ that belongs to $\mathcal{G}_f^{f \cap g}$ is combined with a node from $g$ that belongs to $\mathcal{A}$, line 15 of APPLY expands the node from $f$, and the node (or nodes) from $\mathcal{A}$ remain the same. This will happen until there are no more nodes from $f$ that can be combined with the node (or nodes) from $\mathcal{A}$, and at that point APPLY will simply copy the remaining portion of its output from $\mathcal{G}_g^{aomdd}$. The size of $\mathcal{A}$ is therefore proportional to $\mid \mathcal{G}_g^{f \cap g} \mid$ (because it is the layer of metanodes immediately below $\mathcal{G}_g^{f \cap g}$). A similar argument is valid for the symmetrical case. And there are no combinations between nodes in $\mathcal{G}_g^{aomdd} \setminus \mathcal{G}_g^{f \cap g}$ and $\mathcal{G}_g^{aomdd} \setminus \mathcal{G}_g^{f \cap g}$. The bound follows from all these arguments. □

**Proof of Proposition 5**

The APPLY operation works by constructing the output AOMDD from root to leaves. It first creates a meta-node for the root variable, and then recursively creates its children metanodes by using APPLY on the corresponding children of the input. The worst case that can happen is when the output is not reduced at all, and a recursive call is made for each possible descendant. This corresponds to an unfolding of the full AND/OR search tree based on the context variables, which is exponential in the context size. When the APPLY finishes the context variables, and arrives at the first branching in the bucket pseudo tree, the remaining branches are independent. Similar to the case of OBDDs, where one function occupies a single place in memory, the APPLY can simply create a link to the corresponding branches of the inputs (this is what happens in line 4 in the APPLY algorithm). Therefore, the time and space complexity is at most exponential in the context size. □

**Proof of Theorem 5**

The space complexity is governed by that of **BE**. Since an AOMDD never requires more space than that of a full exponential table (or a tree), it follows that BE-AOMDD only needs space $O(n\, k^{w^*})$. The size of the output AOMDD is also bounded, per layers, by the number of assignments to the context of that layer (namely, by the size of the context minimal AND/OR graph). Therefore, because context size is bounded by treewidth, it follows that the output has size $O(n\, k^{w^*})$. The time complexity follows from Proposition 5, and from the fact that the number of functions in a bucket cannot exceed $r$, the original number of functions. □

**Proof of Proposition 6**

It suffices to prove the proposition for $m = 2$. The general result can then be obtained by induction. It is essential that the function is defined by a constraint network (i.e., the values are only 0 or 1), and that the function takes value 1 at least for one assignment. The value 1 denotes consistent assignments (solutions), while 0 denotes inconsistent assignments. Suppose $f = f_1 \otimes f_2$. Let's denote by $x$ a full assignment to $\mathbf{X}$, and by $x^1$ and $x^2$ the projection of $x$ over $\mathbf{X}^1$ and $\mathbf{X}^2$, respectively. We can write $x = x^1 x^2$ (concatenation of partial assignments). It follows that $f(x) = f_1(x^1) * f_2(x^2)$. Therefore, if $f(x) = 1$, it must be that $f_1(x^1) = 1$ and $f_2(x^2) = 1$. We claim that for any $x^1$, $f_1(x^1) = 1$ only if there exists some $x^2$ such that $f(x^1 x^2) = 1$. Suppose by contradiction that there exist some $x^1$ such that $f_1(x^1) = 1$ and $f(x^1 x^2) = 0$ for any other $x^2$. Since $f$ is not always zero, it follows that $f_2$ is not always zero, and therefore there must be some $x^2$ for which $f_2(x^2) = 1$. This leads to a contradiction, and therefore the functions $f_1$ and $f_2$ are uniquely defined by $f$. $\quad\square$

**Proof of Theorem 6**

The proof is by structural induction over the depth of the pseudo tree $\mathcal{T}$. It follows the canonicity proofs for OBDDs (Bryant, 1986) and MDDs (Srinivasan et al., 1990), but extends them from linear orderings to tree orderings that capture function decomposition according to the pseudo tree $\mathcal{T}$. The depth of $\mathcal{T}$, along each of its paths from root to a leaf, is actually the size of the *dependency set*, or the set of variables on which the value of the function depends. Remember that the AOMDD is an AND/OR graph that is completely reduced. We will use the word function, denoted by $f$, to refer to the universal relation, or its characteristic function, defined by the constraint network.

Assume the depth of $\mathcal{T}$ is 0. This means that the function does not depend on any variable, and must be one of the constants 0 or 1. Suppose the function is the constant 0. Then, it must be that the AOMDD does not contain the terminal meta-node **1**, since all the nodes must be reachable along some path, and it would mean that the function can also evaluate to 1. Suppose the AOMDD contains a nonterminal meta-node, say labeled with $X$, where $X$ can take $k$ different values. It must be that all the $k$ children meta-nodes of $X$ are the terminal meta-node **0**. If there are more than one terminal **0**, then the AOMDD is not completely reduced. If there is only one **0**, it follows that the meta-node labeled with $X$ is redundant. Therefore, from all the above, it follows that the AOMDD representing the constant 0 is made of only the terminal **0**. This is unique, and contains the smallest number of nodes. A similar argument applies for the constant 1.

Now, suppose that the statement of the theorem holds for any constraint network that admits a pseudo tree of depth strictly smaller than $p$, and that we have a constraint network with a pseudo tree of depth equal to $p$, where $p > 0$. Let $X$ be the root of $\mathcal{T}$, having domain $\{x_1, \ldots, x_k\}$. We denote by $f_i$, where $i \in \{1, \ldots, k\}$, the functions defined by the restricted constraint network for $X = x_i$, namely $f_i = f|_{X=x_i}$. Let $Y_1, \ldots, Y_m$ be the children of $X$ in $\mathcal{T}$. Suppose we have two AOMDDs of $f$, denoted by $\mathcal{G}$ and $\mathcal{G}'$. We will show that these two AND/OR graphs are isomorphic.

The functions $f_i$ can be decomposed according to the pseudo tree $\mathcal{T}$ when the root $X$ is removed. This can in fact be a forest of independent pseudo trees (they do not share any variables), rooted by $Y_1, \ldots, Y_m$. Based on Proposition 6, there is a unique decomposition $f_i = f_i^{Y_1} * \ldots * f_i^{Y_m}$, for all $i \in \{1, \ldots, k\}$. Based on the induction hypothesis, each of the function $f_i^{Y_j}$ has a unique AOMDD. In the AND/OR graphs $\mathcal{G}$ and $\mathcal{G}'$, if we look at the subgraphs descending from $X = x_i$, they both are completely reduced and define the same function, $f_i$, therefore there exists an isomorphic mapping $\sigma_i$ between them. Let $v$ be the root metanode of $\mathcal{G}$ and $v'$ the root of $\mathcal{G}'$. We claim that $\mathcal{G}$ and $\mathcal{G}'$ are isomorphic according to the following mapping:

$$\sigma(u) = \begin{cases} v', & if \ u = v; \\ \sigma_i(u), & \text{if } u \text{ is in the subgraph rooted by } \langle X, x_i \rangle. \end{cases}$$

To prove this, we have to show that $\sigma$ is well defined, and that it is an isomorphic mapping.

If a meta-node $u$ in $\mathcal{G}$ is contained in both subgraphs rooted by $\langle X, x_i \rangle$ and $\langle X, x_j \rangle$, Then the AND/OR graphs rooted by $\sigma_i(u)$ and $\sigma_j(u)$ are isomorphic to the one rooted at $u$, and therefore to each other. Since $\mathcal{G}'$ is completely reduced, it does not contain isomorphic subgraphs, and therefore $\sigma_i(u) = \sigma_j(u)$. Therefore $\sigma$ is well defined.

We can now show that $\sigma$ is a bijection. To show that it is one-to-one, assume two distinct meta-nodes $u_1$ and $u_2$ in $\mathcal{G}$, with $\sigma(u_1) = \sigma(u_2)$. Then, the subgraphs rooted by $u_1$ and $u_2$ are isomorphic to the subgraph rooted by $\sigma(u_1)$, and therefore to each other. Since $\mathcal{G}$ is completely reduced, it must be that $u_1 = u_2$. The fact that $\sigma$ is onto and is an isomorphic mapping follows from its definition and from the fact that each $\sigma_i$ is onto and the only new node is the root meta-node. Since the AOMDDs only contain one root meta-node (more than one root would lead to the conclusion that the root meta-nodes are isomorphic and should be merged), we conclude that $\mathcal{G}$ and $\mathcal{G}'$ are isomorphic.

Finally, we can show that among all the AND/OR graphs representing $f$, the AOMDD has minimal number of meta-nodes. Suppose $\mathcal{G}$ is an AND/OR graph that represents $f$, with minimal number of meta-nodes, but without being an AOMDD. Namely, it is not completely reduced. Any reduction rule would transform $\mathcal{G}$ into an AND/OR graph with smaller number of meta-nodes, leading to a contradiction. Therefore, $\mathcal{G}$ must be the unique AOMDD that represents $f$.    $\square$

### Proof of Corollary 1

The proof of Theorem 6 did not rely on the scopes that define the constraint network. As long as the network admits the decomposition induced by the pseudo tree $\mathcal{T}$, the universal function defined by the constraint network will always have the same AOMDD, and therefore any constraint network equivalent to it that admits $\mathcal{T}$ will also have the same AOMDD.    $\square$

### Proof of Theorem 7

The constant that is associated with the root is actually the sum of the weights of all solutions. This can be derived from the definition of the weighted AOMDD. The weights of each meta-node are normalized (they sum to 1), therefore the values computed for each OR node by AND/OR search is always 1 (when the task is computing the sum of all solution weights). Therefore, the constant of the weighted AOMDD is always $\sum_x w(x)$ regardless of the graphical model. We will prove that weighted AOMDDs are canonical for functions that are normalized.

Assume we have two different weighted AOMDDs, denoted by $\mathcal{G}^1$ and $\mathcal{G}^2$, for the same normalized function $f$. Let the root variable be $A$, with the domain $\{a_1, \ldots, a_k\}$. Let $x$ denote a full assignment to all the variables. Similar to the above argument for the root constant, because all the meta-nodes have normalized weights, it follows that $w^1(A, a_1) = w^2(A, a_1) = \sum_{x|A=a_1} f(x)$. The superscript of $w^1$ and $w^2$ indicates the AOMDD, and the summation is over all possible assignments restricted to $A = a_1$. It follows that the root meta-nodes are identical. For each value of the root variable, the restricted functions represented in $\mathcal{G}^1$ and $\mathcal{G}^2$ are identical, and we will recursively apply the same argument as above.

However, for the proof to be complete, we have to discuss the case when a restricted function is decomposed into independent functions, according to the pseudo tree. Suppose there are two independent components, rooted by $B$ and $C$. If one of them is the $\mathbf{0}$ function, it follows that the entire function is $\mathbf{0}$. We will prove that the meta-nodes of $B$ in $\mathcal{G}^1$ and $\mathcal{G}^2$ are identical. If $B$ has only one value $b_1$ extendable to a solution, its weight must be 1 in both meta-nodes, so the meta-nodes are identical. If $B$ has more than one value, suppose without loss of generality that the weights are different for the first value $b_1$, and

$$w^1(B, b_1) > w^2(B, b_1). \tag{1}$$

Since $f \neq \mathbf{0}$, there must be a value $C = c_1$ such that $B = b_1, C = c_1$ can be extended to a full solution. The sum of the weights of all these possible extensions is

$$\sum_{x|B=b_1,C=c_1} f(x) = w^1(B, b_1) * w^1(C, c_1) = w^2(B, b_1) * w^2(C, c_1). \tag{2}$$

From Equations 1 and 2 and the fact that the weight are non-zero, it follows that

$$w^1(C, c_1) < w^2(C, c_1). \tag{3}$$

From Equation 1, the fact that $B$ has more than one value and the fact that the weights of $B$ are normalized, it follows that there should be a value $b_2$ such that

$$w^1(B, b_2) < w^2(B, b_2). \tag{4}$$

From Equations 3 and 4, it follows that

$$w^1(B, b_2) * w^1(C, c_1) < w^2(B, b_2) * w^2(C, c_1). \tag{5}$$

However, both sides of the Equation 5 represent the sum of weights of all solutions when $B = b_2, C = c_1$, namely $\sum_{x|B=b_2,C=c_1} f(x)$, leading to a contradiction. Therefore, it must be that Equation 1 is false. Continuing the same argument for all values of $B$, it follows that the meta-nodes of $B$ are identical, and similarly the meta-nodes of $C$ are identical.

If the decomposition has more than two components, the same argument applies, when $B$ is the first component and $C$ is a meta-variable that combines all the other components.    □

**Proof of Theorem 8**

Consider the well known NP-complete problem of 3-COLORING: Given a graph $G$, is there a 3-coloring of $G$? Namely, can we color its vertices using only three colors, such that any two adjacent vertices have different colors? We will reduce 3-COLORING to the problem of computing the semantic treewidth of a graphical model. Let $H$ be a graph that is 3-colorable, and has a non-trivial semantic treewidth. It is easy to build examples for $H$. If $G$ is 3-colorable, then $G \cup H$ is also 3-colorable and will have a non-trivial semantic treewidth, because adding $G$ will not simplify the task of describing the colorings of $H$. However, if $G$ is not 3-colorable, then $G \cup H$ is also not 3-colorable, and will have a semantic treewidth of zero.    □

**Proof of Proposition 7**

Since AOMDDs are canonical representations of graphical models, it follows that the graphical model for which the actual semantic treewidth is realized will have the same AOMDD as $\mathcal{M}$, and therefore the AOMDD is bounded exponentially in the semantic treewidth.    □